\RequirePackage[OT1]{fontenc}
\documentclass[letterpaper, 10 pt, conference]{ieeeconf}
\usepackage{ifthen}
\newcommand\arxivmode{true}

\usepackage{times}

\IEEEoverridecommandlockouts
\usepackage{multicol}
\usepackage{multirow}
\usepackage{stfloats}
\let\labelindent\relax
\usepackage[shortlabels]{enumitem}
\usepackage{mwe}
\usepackage{bbm}
\usepackage{amsfonts}
\usepackage{amsmath}
\usepackage{amssymb}
\usepackage{balance}
\usepackage[usenames,dvipsnames]{xcolor}
\usepackage[bookmarks=true]{hyperref}
\usepackage{url}
\usepackage{mathtools}
\usepackage{subfig}
\usepackage[binary-units]{siunitx}
\usepackage[ruled,linesnumbered]{algorithm2e}
\usepackage[font=small]{caption}
\usepackage{booktabs}
\usepackage{makecell}
\usepackage{pgfplots}
\usepackage{pgfplotstable}
\usepackage{graphicx}
\usepackage{fancyhdr}

\makeatletter
\let\NAT@parse\undefined
\makeatother
\usepackage[numbers,sort&compress]{natbib}
\usepackage[capitalise]{cleveref}

\newcommand{\realspace}[2][]{\mathbb{R}^{{#2}}}

\newcommand{\imagepair}{\mathcal{I}}

\newcommand{\datasetX}{\mathcal{X}}
\newcommand{\pointcloud}{\mathcal{Z}}

\newcommand{\datasetY}{\mathcal{Y}}
\newcommand{\pointx}{\mathbf{x}}
\newcommand{\pointy}{\mathbf{y}}
\newcommand{\pointz}{\mathbf{z}}

\newcommand{\gmm}{p}
\newcommand{\weight}{\pi}
\newcommand{\gweight}{\tau}

\newcommand{\mean}{\boldsymbol{\mu}}
\newcommand{\gmean}{\boldsymbol{\nu}}

\newcommand{\cov}{\mathbf{\Sigma}}
\newcommand{\gcov}{\mathbf{\Lambda}}

\newcommand{\gaussian}[4][\mathcal{N}]{\ensuremath{ {#1} ({#2} \mid {#3}, {#4})}}

\newcommand{\lsogmm}{\gmm_{L}}
\newcommand{\lindexset}{\mathcal{J}}

\newcommand{\gsogmm}{\gmm_{G}}
\newcommand{\gindexset}{\mathcal{K}}

\newcommand{\findexset}{\mathcal{B}}

\newcommand{\hashtab}{H}
\newcommand{\hashkey}{m}
\newcommand{\hashfunc}{h}
\newcommand{\hashindexset}{\mathcal{A}}
\newcommand{\hashvalue}{\mathbf{q}^{\hashindexset}}
\newcommand{\hashkeyset}{\mathcal{M}}
\newcommand{\hashvalueset}{\mathcal{Q}}
\newcommand{\lthres}{\phi}

\newcommand{\smean}{\mean^{\pointx}}
\newcommand{\numwidth}{N_x}
\newcommand{\numheight}{N_y}
\newcommand{\numdepth}{N_z}
\newcommand{\row}{r}
\newcommand{\col}{c}
\newcommand{\slice}{s}
\newcommand{\morigin}{\mathbf{o}}
\newcommand{\mres}{\alpha}

\newcommand{\bandwidth}{\sigma}

\newcommand{\gtpcld}{\pointcloud^{\text{gt}}}
\newcommand{\prpcld}{\pointcloud^{\text{pr}}}
\newcommand{\leafom}{\boldsymbol{\alpha}_{\text{om}}}
\newcommand{\leafnv}{\boldsymbol{\alpha}_{\text{nv}}}
\newcommand{\leaffcgmm}{\boldsymbol{J}}
\newcommand{\leafsogmm}{\boldsymbol{\sigma}}

\definecolor{darkgray176}{RGB}{176,176,176}
\definecolor{green}{RGB}{0,128,0}
\definecolor{lightgray204}{RGB}{204,204,204}

\usepackage[utf8]{inputenc}
\usepackage{pgfplots}
\usepgfplotslibrary{groupplots,dateplot}
\usetikzlibrary{patterns,shapes.arrows}
\pgfplotsset{compat=newest}

\usepackage{tikz}
\usetikzlibrary{spy,calc}
\usetikzlibrary{external,positioning}
\usepgfplotslibrary{external}
\newcommand{\red}[1]{{\color{red}{#1}}}
\newcommand{\blue}[1]{{\color{black}{#1}}}
\newcommand{\zblue}[1]{{\color{blue}{#1}}}
\newcommand{\green}[1]{{\color{green}{#1}}}

\newcommand{\wennie}[1]{{\color{black}{#1}}}
\newcommand{\kshitij}[1]{{\color{black}{#1}}}

\crefname{figure}{Fig.}{Figs.}
\Crefname{figure}{Figure}{Figures}
\crefname{table}{Tab.}{Tabs.}
\Crefname{table}{Table}{Tables}
\crefname{algorithm}{Alg.}{Algs.}
\Crefname{algorithm}{Algorithm}{Algorithms}
\crefname{section}{Section}{Sections}
\Crefname{section}{Section}{Sections}

\usepackage{epstopdf}
\title{\bf Incremental Multimodal Surface Mapping via Self-Organizing Gaussian Mixture Models}

\author{Kshitij Goel and Wennie Tabib%
\thanks{The authors are with The Robotics Institute, Carnegie Mellon University, Pittsburgh, PA 15213 USA
        {\tt\footnotesize \{kshitij,wtabib\}@cmu.edu}}%
}

\fancypagestyle{firstpagestyle}
{
  \fancyhf{}
  \fancyhead[L]{\footnotesize{%
      \begin{center}
          This paper has been accepted for publication in \emph{IEEE Robotics and Automation Letters}. \\ DOI: \href{https://doi.org/10.1109/LRA.2023.3327670}{10.1109/LRA.2023.3327670} \\
        \end{center}
    }}
  \fancyfoot[L]{\footnotesize{%
        \textcopyright 2023 IEEE. Personal use of this material is
        permitted. Permission from IEEE must be obtained for all other uses,
        in any current or future media, including reprinting/republishing this
        material for advertising or promotional purposes, creating new
        collective works, for resale or redistribution to servers or lists, or
        reuse of any copyrighted component of this work in other works.
  }}

}

\begin{document}
\bstctlcite{IEEEexample:BSTcontrol}
\maketitle
\ifthenelse{\equal{\arxivmode}{true}}
{\thispagestyle{firstpagestyle}}%

\begin{abstract}
This letter describes an incremental multimodal surface mapping
methodology\wennie{, which represents the environment as} \blue{a
continuous probabilistic model.}~\blue{This} \blue{model}
\wennie{enables high-resolution reconstruction while
simultaneously compressing spatial and intensity point cloud data.}
\wennie{The strategy employed in this work utilizes Gaussian mixture
models (GMMs) to represent the environment.  While prior GMM-based
mapping works have developed methodologies to determine the number of
mixture components using information-theoretic techniques, these
approaches either operate on individual sensor observations, making
them unsuitable for incremental mapping, or are not real-time viable,
especially for applications where high-fidelity modeling is required.
To bridge this gap, this letter introduces a spatial
hash map for rapid GMM submap extraction combined with an approach to determine
relevant and redundant data in a point cloud. These contributions
increase computational speed by an order of magnitude compared to
state-of-the-art incremental GMM-based mapping. In addition, the
proposed approach yields a superior tradeoff in map accuracy and size when
compared to state-of-the-art mapping methodologies (both GMM- and not
GMM-based). Evaluations are conducted using both simulated and real-world data.
The software is released open-source to benefit the robotics community.}
\end{abstract}


\IEEEpeerreviewmaketitle

\section{Introduction\label{sec:intro}}
\begin{figure}
\ifthenelse{\equal{\arxivmode}{false}}
{%
  \includegraphics[width=\columnwidth,trim=800 400 750 370,clip]{figures/livingroom1_pr.eps}
}%
{%
  \includegraphics[width=\columnwidth,trim=800 400 750 370,clip]{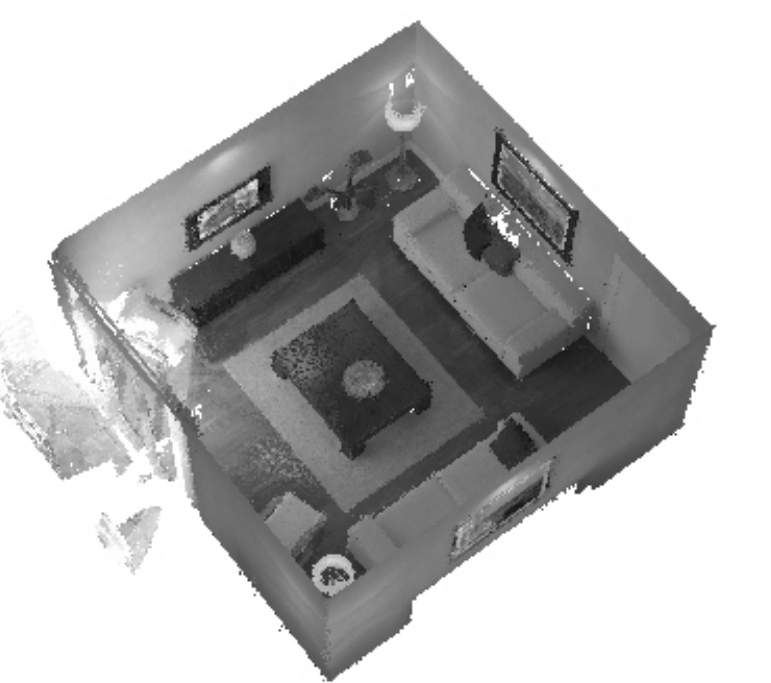}
}%
  \caption{\label{fig:gl-inference}A reconstructed point cloud using spatial and
  intensity information inferred from the compact multimodal point cloud model
  created using the proposed approach. The representation leverages a
  formulation that has been demonstrated to be amenable for higher level robot
  autonomy objectives like exploration in complex, unstructured 3D environments.
  A video is available at: \url{https://youtu.be/VgPEEcbUAnY}.
  }
  \vspace{-0.7cm}
\end{figure}

Robotic exploration systems are being deployed~\blue{to
automate multimodal data collection for applications like artifact
detection~\citep{tranzatto_cerberus_2022}, active thermal
mapping~\citep{tabib_efficient_2016}, and planetary
exploration~\citep{arora_multi-modal_2019}.} For example,
multi-instrument payloads \wennie{(e.g.} range and thermal
sensors\wennie{)} are critical for mapping planetary
caves~\cite{wynne2022fundamental}. \wennie{These} \blue{missions
require an autonomous agent to incrementally create a mathematical
model of the environment using onboard sensors and computers.}
\wennie{The model must accurately represent fine details to enable
operation in close proximity to complex, unknown structures
(e.g. stalactites, thin wires, etc) and be} \wennie{compact to
transfer via} \blue{low-bandwidth communication channels
~\citep{wynne2022fundamental}.} \wennie{Prior work} \blue{has
demonstrated} \wennie{that the exploration performance of a
multi-agent team is impacted by the compactness of the environment
model used in communications restricted environments due to the ability
(or lack thereof) to share information~\citep{goel_rapid_2021,ebadi_lamp_2020}}.

\blue{Few} \wennie{environment models are compact while
enabling} \blue{high-resolution reconstruction and safe
navigation. Octomap~\citep{hornung_octomap_2013},
Voxblox~\citep{oleynikova_voxblox_2017}, and GMM
maps~\citep{tabib_real-time_2019} are the key methodologies that have
been used recently for communication-constrained exploration
by~\citet{agha_nebula_2022},~\citet{tranzatto_cerberus_2022},
and~\citet{goel_rapid_2021}\wennie{,} respectively.} \wennie{These}
\blue{methods enable safe navigation}
\wennie{but suffer from the same limitation in terms of fixing
the \kshitij{highest achievable} map fidelity throughout exploration (e.g., minimum leaf
size for Octomap, voxel size for Voxblox, and number of mixture
components for GMM maps). This is inefficient \kshitij{because
all parts of the environment are assigned the same level of highest fidelity,
leading to larger map sizes and communication inefficiency~\cite{ebadi_lamp_2020}.}
}

\wennie{To enable large-scale exploration with many robots
simultaneously sharing information, the mapping algorithm must adapt
\kshitij{to the scene complexity}.}
\wennie{Our} \blue{recent work~\citep{goel_probabilistic_2023}, the
Self-Organizing Gaussian Mixture Models (SOGMMs)}\wennie{, proposes
an information-theoretic approach for automatically determining the number
of mixture components from the underlying sensor data; however, the approach
operates on single observations. Because consecutive sensor observations have
significant overlap, the formulation in~\citep{goel_probabilistic_2023}
is not well-suited for incremental mapping.}

\wennie{Prior
GMM mapping approaches not demonstrated onboard a robot} \blue{either require
iterating over the mixture
components~\citep{srivastava_efficient_2017} or use approximate geometric
projection
methods~\citep{dhawale_efficient_2020,tabib_autonomous_2021}.  The
computational requirements for these methods increase prohibitively
with the number of components in the GMM map.} \wennie{To bridge these gaps,
we extend the approach of~\citep{goel_probabilistic_2023} and
provide the following novel contributions (\cref{fig:gl-inference}):}
\begin{enumerate}
\item \wennie{a methodology which alleviates the computational burden
      of submap extraction by innovating a spatial hash table of mixture components;}
\item \kshitij{an efficient method to incrementally update the global environment model
      leveraging the log-likelihood of the point cloud data with respect to the extracted submap;}
\item \wennie{extensive evaluations on simulated and real-world datasets with comparisons
      against state-of-the-art approaches (both GMM-based and not GMM-based); and}
\item \wennie{an} \blue{open-source release of the approach for the
      benefit of robotics research and industry.}
\end{enumerate}

The letter \wennie{proceeds with an overview of related work
(\cref{sec:related_work}). \Cref{sec:approach} presents the
proposed approach and \cref{sec:results} details the
results.} \blue{Limitations of the approach} \wennie{are detailed
in~\cref{sec:limitations} and the letter is concluded with
\cref{sec:conclusion}.}


\section{Related Work\label{sec:related_work}}
\wennie{This section} reviews mapping
methods that enable high-\kshitij{resolution} multimodal
reconstruction~\wennie{and occupancy modeling for} 3D robotic
exploration.

Early exploration works leveraged discrete occupancy grid
maps~\cite{elfes_using_1989} \wennie{for rapid volumetric
querying}~\citep{yamauchi_frontier-based_1998,burgard_collaborative_2000}.
However, \wennie{voxel grids} are subject to aliasing due to the independence
assumption between cells (voxels) leading to poor multimodal
reconstruction. To alleviate the effects of aliasing,
smaller voxel sizes can be used\wennie{; however,} the
memory footprint substantially increases as the number of voxels
scales cubically according to the environment extents.

To mitigate the memory challenges,~\citet{hornung_octomap_2013}
leverages octrees to enable a multi-resolution and hierarchical
volumetric representation. This method has been used by recent
exploration frameworks due to its scalability improvements over the
occupancy grids at small voxel
sizes~\citep{zhang_fsmi_2020,tranzatto_cerberus_2022}.  However, the
insertion cost for new point clouds into Octomap and the cost of voxel
state access (for the purposes of informative planning) is higher than
occupancy grids. To decrease \kshitij{these} costs while
\kshitij{leveraging} a hierarchical approach, the OpenVDB method
by~\citet{museth_vdb_2013} utilizes B+ trees,~\wennie{which yields}
increased efficiency in frontier extraction at large spatial
scales~\citep{scherer_resilient_2022}. An alternate approach
by~\citet{oleynikova_voxblox_2017} called Voxblox enables superior,
constant-time, insertion and queries to the voxel grid while enabling
informative planning and multimodal reconstruction
capabilities~\citep{tranzatto_cerberus_2022}. Voxblox uses a regular
occupancy grid \wennie{and} Truncated Signed Distance Fields (TSDFs)
to approximately alleviate the aliasing due to the discrete
representation. However, \kshitij{in terms of communication efficiency}
the \wennie{representation suffers from the same drawbacks as occupancy grids}.
In this work, we present a mapping method that achieves higher multimodal
reconstruction accuracy than Voxblox and Octomap through compact and continuous
point cloud models.

\wennie{\citet{saarinen_3d_2013} motivate the development of NDTMap,
which uses a Gaussian density in each cell, by arguing that larger
voxels may be used since} \blue{the Gaussian density better
approximates the surface geometry.} However, this representation also
suffers from the aliasing challenges of a regular occupancy grid as
each Gaussian density is considered an independent component of a
uniformly-weighted Gaussian mixture model
(GMM).\wennie{~\citep{corah_communication-efficient_2019,tabib_real-time_2019}
relax the assumption of uniform weights, by using a}
maximum-likelihood fit over the point cloud data to create a global
map that is represented as a GMM without the use of a discrete
grid. However, these works require specifying the number of mixture
components before operation which limits the maximum achievable
fidelity of the map. We bridge
this gap by proposing a GMM-based approach that enables creating a
map representation that \kshitij{increases the} model fidelity
incrementally via an information-theoretic self-organizing
approach~\citep{goel_probabilistic_2023} and enables scalable and
efficient inference via spatial hashing.

\blue{Neural Radiance Fields (NeRFs)~\citep{mildenhall_nerf_2020}}
\wennie{enable photorealistic environment rendering at lower memory
costs; however, incremental mapping with implicit representations
are known to suffer from catastrophic
forgetting~\cite{zhong_shine-mapping_2023-1}}.
\wennie{Catastrophic forgetting is the problem of forgetting old
knowledge after training with new data~\cite{mccloskey1989catastrophic}.  To
mitigate this issue, some incremental NeRF mapping
approaches~\cite{imap,isdf} retain keyframes from historical data and
replay them with current data to train the network; however, this
approach requires more memory to store the
keyframes~\cite{zhong_shine-mapping_2023-1}.}\wennie{~\citet{zhong_shine-mapping_2023-1}
develop a technique for large scale Signed Distance Field (SDF) mapping using
LiDAR, but it is not clear how robust it will be to catastrophic forgetting
when intensity data is incorporated. In contrast, \kshitij{the proposed approach
adaptively increases the fidelity of the parametric GMM-based environment model depending on the scene
complexity. This way, while offering an implicit representation of the surface, no special
consideration for catastrophic forgetting is required.}}

\blue{
Finally, note that the
current state-of-the-art in radiance field rendering in the computer
graphics literature leverages 3D Gaussian densities in place of neural
networks~\citep{keselman_approximate_2022,kerbl_3d_2023}, demonstrating superior
performance in both training and inference compared to NeRFs. Our method uses a
mixture of 4D Gaussian densities to jointly model intensity and spatial data.
Therefore, incorporating radiance field rendering within the proposed
incremental mapping method is an exciting direction for future work.}


\section{Approach\label{sec:approach}}
\begin{figure*}
  \includegraphics[width=\textwidth,trim=0 825 400 0,clip]{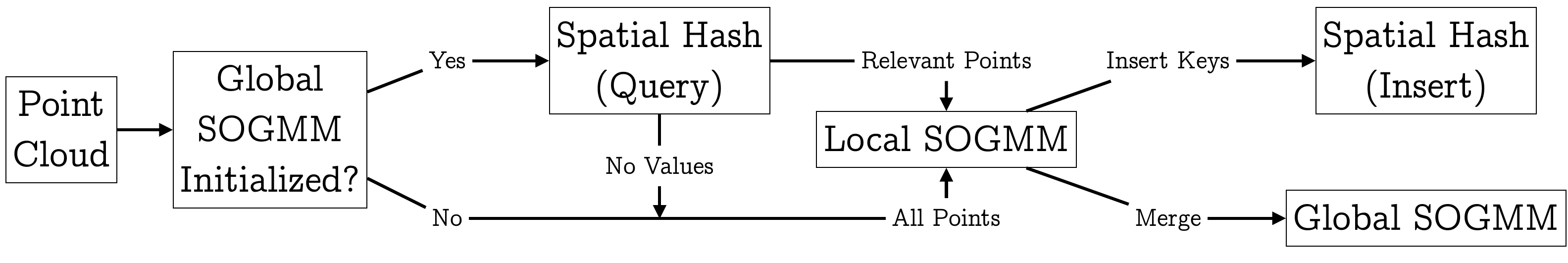}
  \caption{\label{fig:surf-map-flow}Information flow during surface point cloud modeling
  via the proposed incremental mapping approach (\cref{ssec:proposed-mapping}).}
  \vspace{-0.7cm}
\end{figure*}

This section provides details of the~\wennie{mapping methodology}. The
incremental mapping methodology is presented
in~\cref{ssec:proposed-mapping} and~\cref{ssec:proposed-inference}
describes the inference method used for multimodal surface
reconstruction.

\textbf{Notation.} We follow the notation from~\cite{goel_probabilistic_2023},
which uses small letters for scalars (e.g., $x$, $y$), bolded small letters for
vectors (e.g., $\pointx$, $\pointy$), bolded capital letters for matrices (e.g.,
$\mathbf{X}$, $\mathbf{Y}$), and calligraphic letters for sets (e.g.,
$\imagepair$, $\datasetX$, $\datasetY$).

%

\subsection{Incremental Multimodal Surface Mapping\label{ssec:proposed-mapping}}
Before describing the algorithm, \wennie{three key data structures
are described}: the Local SOGMM, Global SOGMM, and Spatial
Hash \blue{table}.

\textbf{Local SOGMM.} A GMM model created via the SOGMM method using the relevant points corresponding to
the latest multimodal point cloud. Each point $\pointz$ in this point cloud is
assumed to be of the form, $\pointz = \{ (\pointx, i) \mid \pointx \in \realspace{3}, i \in [0.0, 1.0]\}$.
Formally, the GMM model is represented as the function
  $\lsogmm \equiv \lsogmm(\pointz) = \sum_{j \in \lindexset} \weight_j \gaussian{\pointz}{\mean_j}{\cov_j}$,
where $\weight_j$, $\mean_j$, and $\cov_j$ are the weight, mean, and covariance
for the mixture component associated with index $j$ in $\lsogmm$.  Each mixture
component is a Gaussian probability density
$\gaussian{\pointz}{\mean_j}{\cov_j}$. The set of indices is denoted by
$\lindexset$. The sum of weights \blue{must} be 1, $\sum_{j \in \lindexset}
\weight_j = 1$, for a valid $\lsogmm$.

\textbf{Global SOGMM.} A GMM model created after merging all prior $\lsogmm$
models. \wennie{T}he global model contains $|\gindexset| \geq |\lindexset|$ mixture components. Formally,
  $\gsogmm \equiv \gsogmm(\pointz) = \sum_{k \in \gindexset} \gweight_k \gaussian{\pointz}{\gmean_k}{\gcov_k}$,
where different index and parameter symbols are used to \blue{notationally} differentiate $\gsogmm$ from $\lsogmm$.
Similar to $\lsogmm$, the set of indices is denoted by $\gindexset$ and $\sum_{k \in \gindexset} \weight_k = 1$
must hold for a valid $\gsogmm$.

\textbf{Spatial Hash.} A hash table $\hashtab : \hashkeyset \to \hashvalueset,
\hashkey \mapsto \hashvalue[\hashfunc(\hashkey)]$ is created to map any point \wennie{in
3D space} (key $\hashkey \in \hashkeyset$) to a vector of mixture
component indices in $\gsogmm$ (value $\hashvalue \in \hashvalueset$) that are
within a fixed volume around the point. \blue{This fixed volume is a cube with side
length $\mres$.} The hash function $h : \hashkeyset \to \hashindexset$ maps the
keys to an index set $\hashindexset = \{ 0, 1, \ldots, |\hashindexset| - 1 \}$
used to insert into and query from $\hashtab$.

\Cref{fig:surf-map-flow} \wennie{provides} an overview of the multimodal surface
mapping method. There are three steps: (1) creating $\lsogmm$, (2) merging
$\lsogmm$ into $\gsogmm$, and (3) spatially hashing $\lsogmm$ into $\hashtab$.
\wennie{Details of each step are provided in the following sections.}

\subsubsection{Creating $\lsogmm$\label{sssec:local-sogmm}}
For each point cloud $\pointcloud$, the relevant subset of the point
cloud $\pointcloud^r$ \wennie{is determined}.  If the number of points
in this set is greater than a pre-specified threshold, then $\lsogmm$
is created\wennie{;} otherwise\wennie{,} $\pointcloud^r$ is cached and
used along with the subsequent frames.  \wennie{$\pointcloud^r$ represents
points} not already modeled by $\gsogmm$. Therefore, if $\gsogmm$ is
not initialized, all points are treated as relevant ($\pointcloud^r =
\pointcloud$). Otherwise, $\pointcloud^r$ is determined using a
threshold ($\lthres$) on the log-likelihood\wennie{~\citep{srivastava_efficient_2017}} that points $\pointcloud$
originated from the model $\gsogmm$,
\begin{align}
\pointcloud^r = \{ \pointz \in \pointcloud \mid \mathcal{L}(\pointz) = \ln{\left( \gsogmm(\pointz) \right)}, \mathcal{L}(\pointz) < \lthres \}. \label{eq:zr-4d}
\end{align}
\blue{However, this approach has two
drawbacks when used with multimodal point clouds. First, thresholding the
log-likelihood scores via~\cref{eq:zr-4d} over the multimodal point cloud directly does not yield
the intended $\pointcloud^r$ as the $4^{\text{th}}$ dimension contains intensity
data, which is not in the metric space of the other three dimensions.}  Second,
as the size of the model $\gsogmm$
(i.e., the value of $K$) increases over time, the time complexity of
calculating~\cref{eq:zr-4d} increases linearly. Performing this computation for
all points in $\pointcloud$ is prohibitive for real-time operation on
computationally-constrained robotic systems.

\begin{figure*}
  \centering
  \ifthenelse{\equal{\arxivmode}{false}}
  {%
  \subfloat[\label{sfig:given-pclds}]{\includegraphics[height=5.65cm,trim=70 70 20 110,clip]{figures/given-pclds.eps}}\hfill%
  \subfloat[\label{sfig:4d-check}]{\includegraphics[height=5.65cm,trim=70 70 250 110,clip]{figures/4d-check.eps}}\hfill%
  \subfloat[\label{sfig:3d-check}]{\includegraphics[height=5.65cm,trim=70 70 250 110,clip]{figures/3d-check.eps}}\hfill%
  \subfloat[\label{sfig:3d-check-fov}]{\includegraphics[height=5.65cm,trim=70 70 250 110,clip]{figures/3d-check-fov.eps}}%
  }%
  {%
  \subfloat[\label{sfig:given-pclds}]{\includegraphics[height=5.65cm,trim=70 70 20 110,clip]{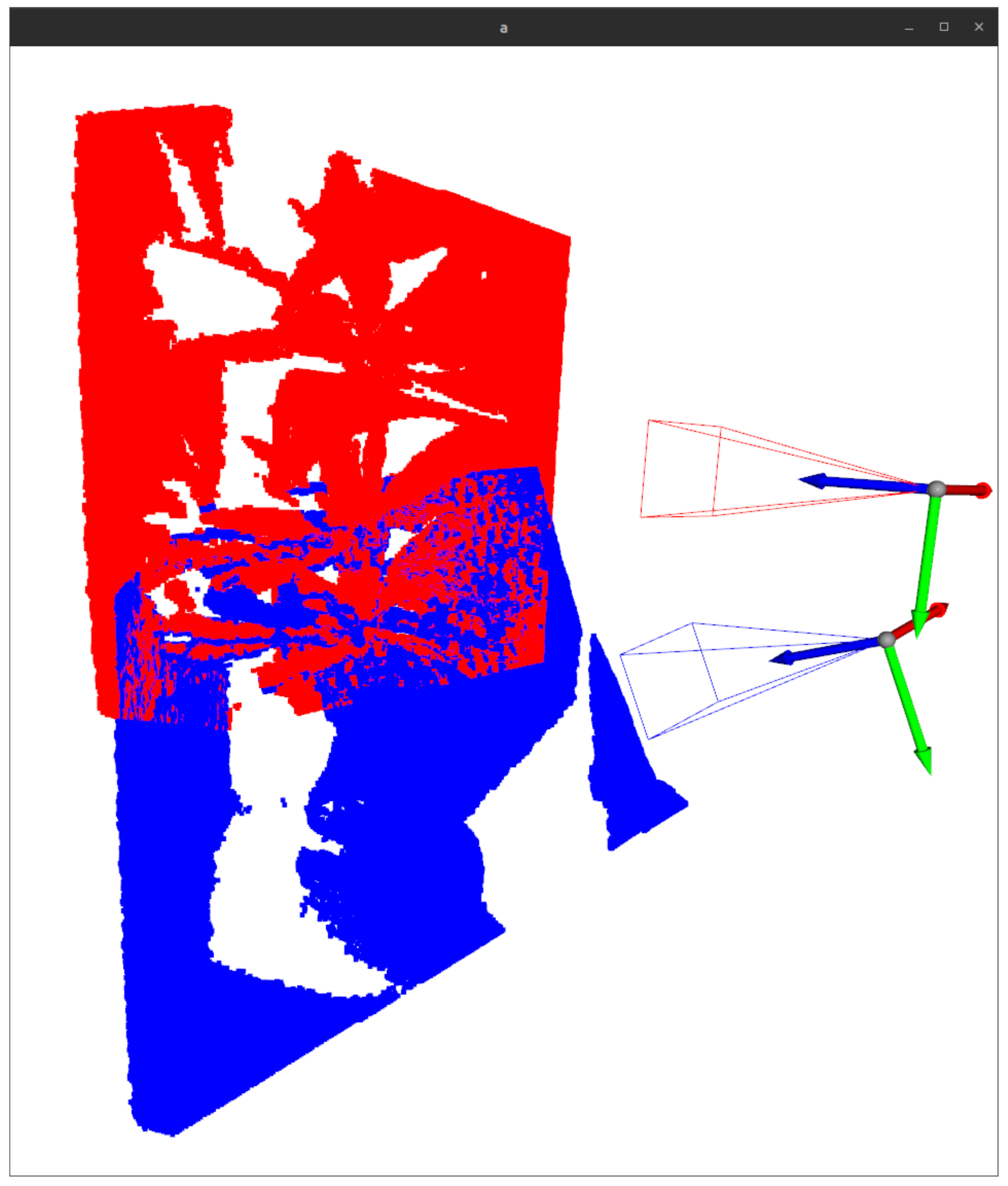}}\hfill%
  \subfloat[\label{sfig:4d-check}]{\includegraphics[height=5.65cm,trim=70 70 250 110,clip]{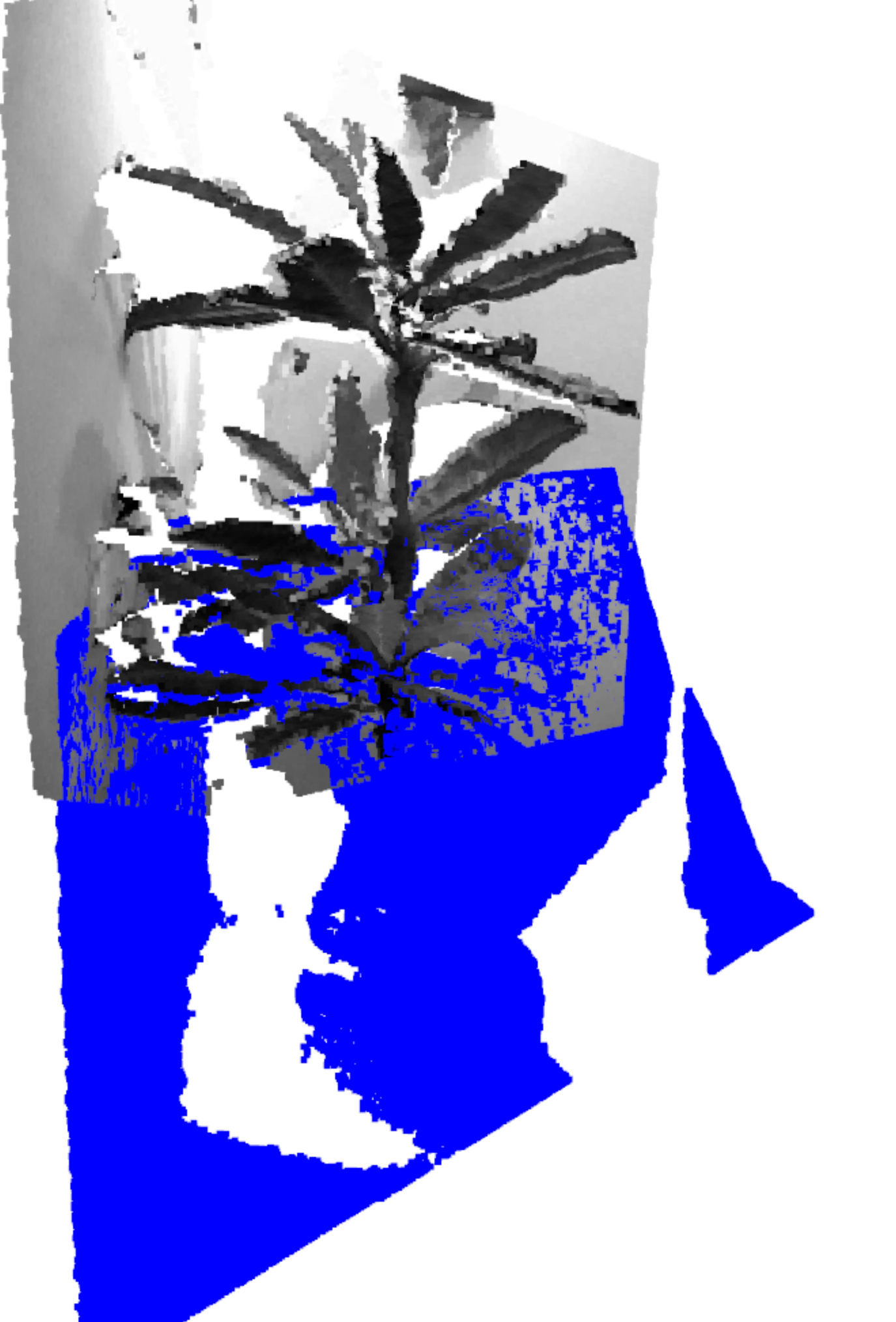}}\hfill%
  \subfloat[\label{sfig:3d-check}]{\includegraphics[height=5.65cm,trim=70 70 250 110,clip]{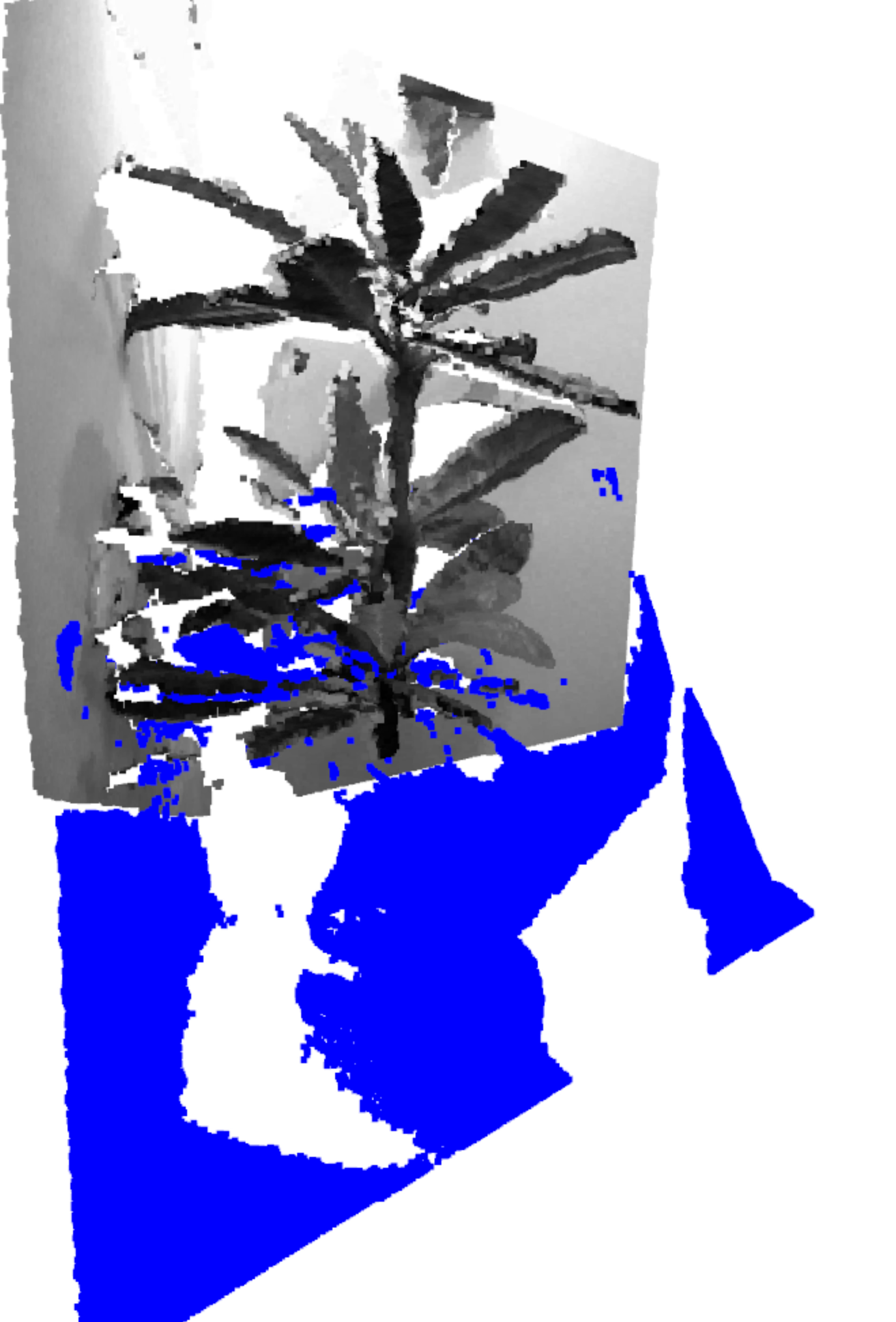}}\hfill%
  \subfloat[\label{sfig:3d-check-fov}]{\includegraphics[height=5.65cm,trim=70 70 250 110,clip]{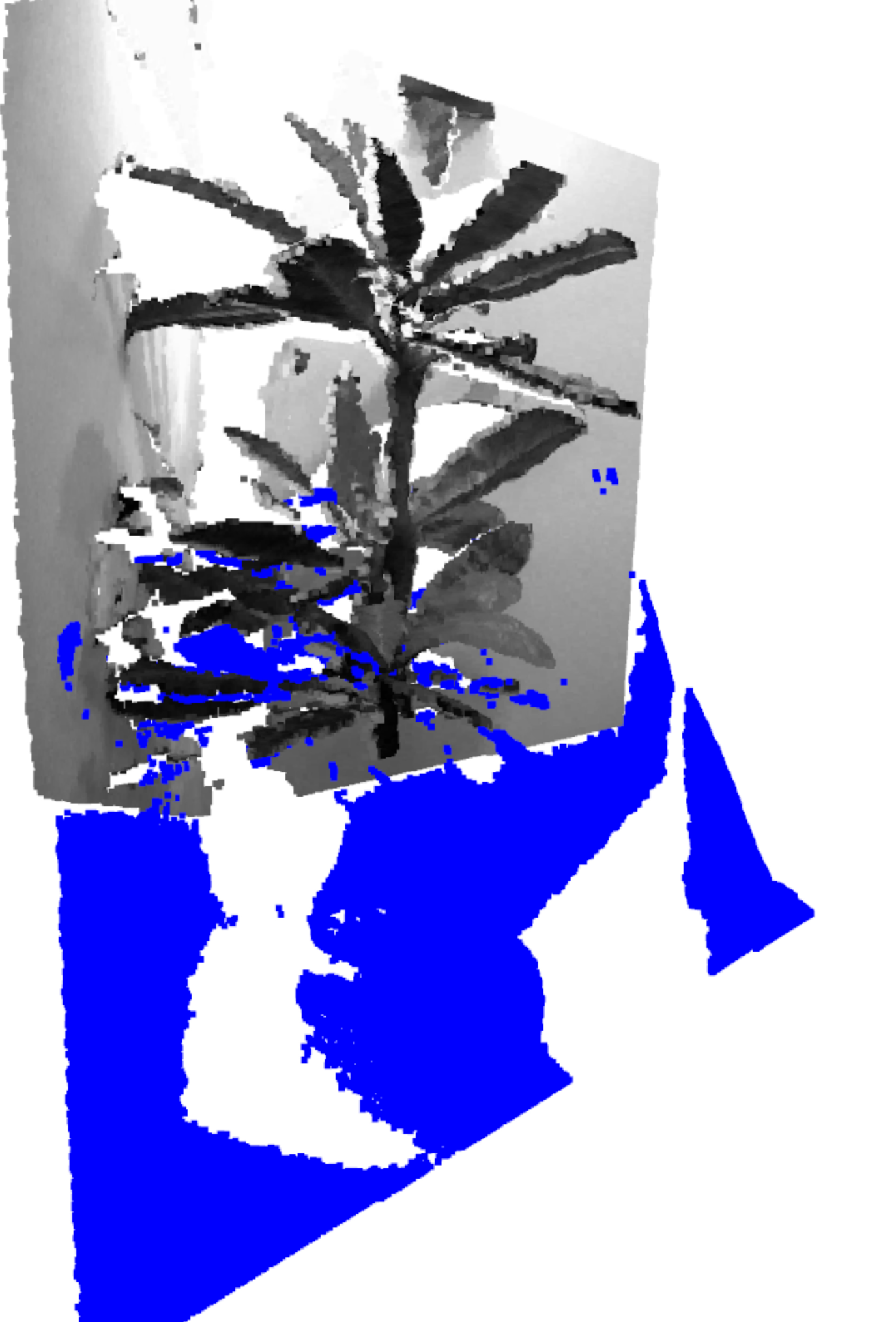}}%
  }%
  \caption{\label{fig:zr-marginal}Illustration of the relevant point cloud calculation using two multimodal point
  clouds, \red{$\pointcloud_1$} and \zblue{$\pointcloud_2$} (\cref{sssec:local-sogmm}). The objective is to find the relevant
  point cloud, \zblue{$\pointcloud^r_2$}, \wennie{from} \zblue{$\pointcloud_2$} using $\gsogmm$\wennie{, which is} created from \red{$\pointcloud_1$}.
  \protect\subref{sfig:given-pclds} shows the 3D parts of these point clouds in
  different colors and the associated 3D poses.~\protect\subref{sfig:4d-check} shows
  \zblue{$\pointcloud^r_2$} and \red{$\pointcloud_1$} with intensity values, when~\cref{eq:zr-4d} is used.
  \protect\subref{sfig:3d-check} shows the same but when~\cref{eq:zr-3d} is used.
  Notice that in the former case \zblue{$\pointcloud^r_2$} contains more misclassified points
  that overlap with $\gsogmm$ than \wennie{in} the latter case.~\protect\subref{sfig:3d-check-fov} shows the output
  \zblue{$\pointcloud^r_2$} when only a subset ($|\findexset| = 480$) of components in $\gsogmm$ ($|\gindexset| = 1165$)
  derived using the hash table $\hashtab$ are used. This output is similar to~\protect\subref{sfig:3d-check}.
  \wennie{The p}oint clouds \wennie{are sourced from} from the \blue{real-world} Lounge dataset~\citep{zhou_dense_2013}.
  \textit{This figure is best viewed in color.}
  }
  \vspace{-0.7cm}
\end{figure*}

The first challenge is addressed by utilizing the marginal probability density
$p(\pointx)$ instead of $p(\pointz)$ for the log-likelihood calculation, i.e.,
\begin{align}
\pointcloud^r &= \{ \pointz \in \pointcloud \mid \pointz = (\pointx, i), \mathcal{L}(\pointx) < \lthres \} \text{ where,}  \label{eq:zr-3d}\\
\mathcal{L}(\pointx) &= \ln{ \left( \sum_{k \in \gindexset} \gweight_k \gaussian{\pointx}{\gmean^{\pointx}_k}{\gcov^{\pointx\pointx}_k} \right)}, \label{eq:zr-3d-compute}\\
\gmean_k &= [\gmean^{\pointx}_k, \gmean^{i}_k]^{\top},\text{ and } \gcov_k = \begin{bmatrix} \gcov^{\pointx\pointx}_k & \gcov^{\pointx i}_k\\ \gcov^{i \pointx}_k & \gcov^{ii}_k \end{bmatrix}. \label{eq:global-marg}
\end{align}
\wennie{\Cref{fig:zr-marginal} details the effect of
using~\cref{eq:zr-3d} instead of~\cref{eq:zr-4d}.}
\blue{The value of $\lthres$ is determined empirically, as done
in~\citep{srivastava_efficient_2019}, and it is fixed for the
synthetic and real-world scenarios.}

The second challenge is addressed by selecting only the subset of mixture
components from $\gsogmm$ (i.e., selecting $\findexset \subseteq \gindexset$)
that are overlapped by or close to the points in $\pointcloud$. This way we can
reduce the number of summands in~\cref{eq:zr-3d-compute}.  The hash table
$\hashtab$ is leveraged for this purpose. Each 3D point $\pointx$ from $\pointz
\in \pointcloud$ is a key $\hashkey$ for the hash function $\hashfunc$ that is
used to search $\hashtab$ for the closest vector of mixture component indices,
$\hashvalue[\hashfunc(\hashkey)]$. After attaining these vectors for all points
in $\pointcloud$, the unique set of mixture component indices form the index set
$\findexset$. For the example scenario
in~\cref{fig:zr-marginal},~\cref{sfig:3d-check-fov} shows the output after this
approximation. The output is similar to the case when the original set of
components $\gindexset$ (\cref{sfig:3d-check}) is used but $\findexset$ is
smaller ($480$ elements instead of $1165$). Consequently, for this example the
time taken to compute~\cref{eq:zr-3d-compute} with $\gindexset$ is
$\SI{0.35}{\second}$ whereas with $\findexset$ it is $\SI{0.25}{\second}$
($28\%$ faster).  Note that the ratio $|\gindexset| / |\findexset|$ grows over
time as the size of $\gsogmm$ increases when point clouds from new regions are
observed. \wennie{An analysis of the computational savings using this approach
is provided in~\cref{ssec:ll-compare}.}

\subsubsection{Merging $\lsogmm$ into $\gsogmm$\label{sssec:global-sogmm}}
After the $\lsogmm$ model is created using $\pointcloud^r$, it is merged with
the global point cloud model $\gsogmm$ by appending the parameters and
re-normalizing the weights. Let the global model before merging be $\gsogmm^{t}$
and after merging be $\gsogmm^{t+1}$. The parameters for $\gsogmm^{t+1}$ are
given by
  $\boldsymbol{\gweight}^{t+1} = [\boldsymbol{\gweight}^{t}, \boldsymbol{\weight}]^{\top}$ such that $\sum_{b \in \findexset} \gweight_b^{t+1} = 1$,
  $\gmean^{t+1} = [\gmean^{t}, \mean]^{\top}$, and $\gcov^{t+1} = [\gcov^{t}, \cov]^{\top}$.
The index set for the global model is also updated and the number of components increase
accordingly, $| \gindexset |^{t+1} = | \gindexset |^{t} + | \lindexset |$.


\subsubsection{Spatially Hashing $\lsogmm$ into $\hashtab$\label{sssec:spatial-hash-global}}
In addition to the global model, the hash table $\hashtab$ is updated using the mixture components from
the latest local model $\lsogmm$. A total of $|\lindexset|$ hash keys are inserted into the table where each hash
key $\hashkey_j$ is the spatial part of the mean position $\smean_j$ along with
points generated at constant probability ellipsoids corresponding to $68\%$ ($1$-sigma),
$95\%$ ($2$-sigma) and $99.7\%$ ($3$-sigma) of the data points,
for all $j \in \lindexset$.

The hash function $\hashfunc$ that maps $\hashkey_j$ to an index in $\hashindexset$ is given by
$\hashfunc(\hashkey_j) \equiv \hashfunc(\smean_j) = \numdepth(\row(\smean_j) \numwidth + \col(\smean_j)) + \slice(\smean_j)$
such that, $\row(\smean_j) = \lfloor (\mean^y_j - \morigin^y) / \mres \rfloor$,
$\col(\smean_j) = \lfloor (\mean^x_j - \morigin^x) / \mres \rfloor$,
and $\slice(\smean_j) = \lfloor (\mean^z_j - \morigin^z) / \mres \rfloor$.
Here, $[\numwidth, \numheight, \numdepth]$ are the number of cells along each
axis of a 3D regular grid of spatial resolution $\mres$,
$\morigin = -\frac{1}{2}\mres [\numwidth, \numheight, \numdepth]$ is the origin position of
this grid, and $\lfloor . \rfloor$ is the floor operator. Intuitively, the hash
function $\hashfunc$ assigns the mean positions $\smean$ into a sparse grid of
a pre-specified extent $[\numwidth, \numheight, \numdepth]$ and resolution
$\mres$.

The value corresponding to each key is the index of the component $j$ in the
global model after the merging step (\cref{sssec:global-sogmm}) is complete.
Thus, the value corresponding to the hash key $\hashkey_j$ is $| \gindexset
|^{t} + j$. It is possible that multiple hash keys are mapped to the same cell
in the grid (i.e., hash collisions are possible). An example scenario is when
$\mres$ is large and a subset of means $\smean$ are spatially within $\mres$
distance. In this case, we want to store all the values in a vector.  This is
why the value set $\hashvalueset$ is defined as a set of vectors as opposed to a
set of integers. If a hash collision occurs for any two keys $\hashkey_f$ and
$\hashkey_g$ (i.e., $\hashfunc(\hashkey_f) = \hashfunc(\hashkey_g)$), the values
are appended into the vector $\hashvalue[\hashfunc(\hashkey_f)]$.

\subsection{Global Spatial and Intensity Inference\label{ssec:proposed-inference}}
Given the global model $\gsogmm$, we want to reconstruct the environment
spatially along with the intensity values. The marginal global model given by
$\boldsymbol{\gweight}$, $\gmean^{\pointx}$, and $\gcov^{\pointx \pointx}$ (as defined
by~\cref{eq:global-marg}) is used for spatial inference and densely sampled
using the Box-Muller transform~\citep{box1958note}. The conditional probability
density $p(i \mid \pointx)$ (as noted in~\citep{goel_probabilistic_2023}), is
used to infer intensity at the sampled spatial points. This inference is performed
in batches of components from $\gsogmm$. The batch size is determined based on
the available memory on the CPU used to perform inference.
An example of the reconstruction obtained for the Living Room
dataset~\citep{choi_robust_2015} is shown in~\cref{fig:gl-inference}.


\section{Results\label{sec:results}}
\blue{The experimental results are divided into two parts. In~\cref{ssec:ll-compare}, the
computational performance gain offered by the proposed incremental mapping approach
due to the spatial hash formulation is compared with the prior work on GMM-based
multimodal mapping~\citep{srivastava_efficient_2017}. In~\cref{ssec:gl-map-compare}
the reconstruction accuracy from the global map created through the proposed approach
is compared with Octomap~\citep{hornung_octomap_2013}, Voxblox
(Nvblox\footnote{Nvblox is the GPU-accelerated extension of Voxblox:
\url{https://github.com/nvidia-isaac/nvblox}})~\citep{oleynikova_voxblox_2017},
and GMM-based maps that use a fixed number of components
(FCGMM)~\citep{tabib_autonomous_2021}. These methods are chosen as baselines
because they have been used for our target application, multi-robot 3D
exploration. We will use the ``Method-Parameter'' notation to denote
the parameter being used. For example, Proposed-$0.02$ denotes the proposed
approach with the bandwidth parameter $\sigma = 0.02$.}

\begin{figure}
  \centering
\blue{
  \ifthenelse{\equal{\arxivmode}{false}}
  {%
  \subfloat[Baseline~\citep{srivastava_efficient_2017} with varying $|\lindexset|$\label{sfig:ll-base}]{\input{figures/ll_baseline_cpu.tex}}%
  \subfloat[Proposed with varying $\bandwidth$\label{sfig:ll-isogmm}]{\input{figures/ll_isogmm.tex}}\\
  }%
  {%
  \subfloat[Baseline~\citep{srivastava_efficient_2017} with varying $|\lindexset|$\label{sfig:ll-base}]{\includegraphics{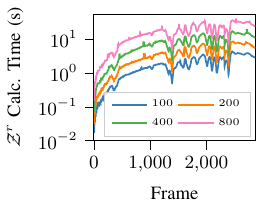}}%
  \subfloat[Proposed with varying $\bandwidth$\label{sfig:ll-isogmm}]{\includegraphics{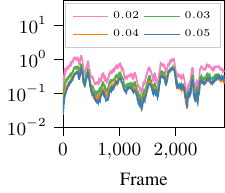}}\\
  }%
  \begin{minipage}{0.55\columnwidth}
  {\footnotesize \subfloat[Cumulative time taken\label{sfig:ll-times}]{\begin{tabular}{lcc}
\toprule
Method & Time (s) & $\Delta$\\
\midrule
FCGMM-$800$ & 44751.6 & -- \\
FCGMM-$400$ & 20702.3 & -- \\
FCGMM-$200$ & 9957.6 & -- \\
FCGMM-$100$ & 4774.7 & -- \\
\hline
Proposed-$0.02$ & 1239.6 & $\mathbf{36.1\times}$\\
Proposed-$0.03$ & 666.1 & $\mathbf{31.0\times}$\\
Proposed-$0.04$ & 488.0 & $\mathbf{20.4\times}$\\
Proposed-$0.05$ & 436.9 & $\mathbf{10.9\times}$\\
\bottomrule
\end{tabular}}}
  \end{minipage}%
  \begin{minipage}{0.45\columnwidth}
  \ifthenelse{\equal{\arxivmode}{false}}
  {%
    \subfloat[Ablation for $\mres$\label{sfig:ll-sp-abl}]{\input{figures/ll_sp_abl.tex}}
  }%
  {%
    \subfloat[Ablation for $\mres$\label{sfig:ll-sp-abl}]{\includegraphics{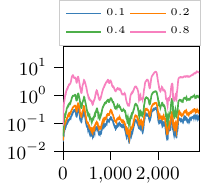}}
  }%
  \end{minipage}%
}
  \caption{\label{fig:ll-comparison}\blue{
  Comparison of the relevant subset $\pointcloud^r$ calculation time between
  the prior work on multimodal GMM mapping~\citep{srivastava_efficient_2017} and
  the proposed approach. The per-frame calculation time in seconds is plotted
  for~\protect\subref{sfig:ll-base} different values of fixed number\wennie{s} of
  components $|\lindexset|$ and~\protect\subref{sfig:ll-isogmm} different values
  of the bandwidth parameter $\bandwidth$ for the proposed
  method.~\protect\subref{sfig:ll-times} Notice that the spatial hash
  (\cref{sssec:spatial-hash-global}) enables an order of magnitude improvement
  and that the performance gains increase monotonically with model size.
  \protect\subref{sfig:ll-sp-abl} shows an ablation of calculation times for
  different values of the spatial hash resolution parameter $\mres$.
  }
  \vspace{-0.7cm}
}
\end{figure}

\begin{figure*}
  \centering
  \blue{
  \ifthenelse{\equal{\arxivmode}{false}}
  {%
  \subfloat[MRE (lower is better)\label{sfig:lr-mre}]{\input{figures/mre_livingroom1_rebuttal.tex}}%
  \subfloat[Precision (higher is better)\label{sfig:lr-prec}]{\input{figures/precision_livingroom1_rebuttal.tex}}%
  \subfloat[Recall (higher is better)\label{sfig:lr-rec}]{\input{figures/recall_livingroom1_rebuttal.tex}}%
  \subfloat[PSNR (higher is better)\label{sfig:lr-psnr}]{\input{figures/psnr_livingroom1_rebuttal.tex}}
  }%
  {%
  \subfloat[MRE (lower is better)\label{sfig:lr-mre}]{\includegraphics{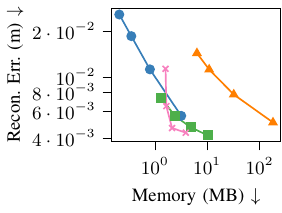}}%
  \subfloat[Precision (higher is better)\label{sfig:lr-prec}]{\includegraphics{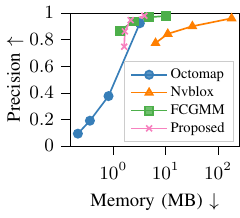}}%
  \subfloat[Recall (higher is better)\label{sfig:lr-rec}]{\includegraphics{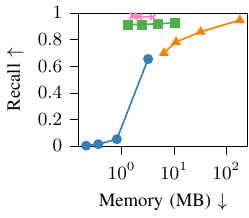}}%
  \subfloat[PSNR (higher is better)\label{sfig:lr-psnr}]{\includegraphics{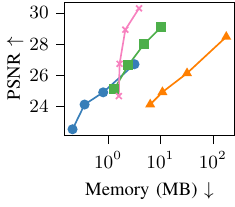}}
  }%
  }
  \caption{\label{fig:livingroom1-plot}
    \blue{Quantitative comparison of}\wennie{~\protect\subref{sfig:lr-mre} reconstruction error,
      ~\protect\subref{sfig:lr-prec} precision,~\protect\subref{sfig:lr-rec} recall, and
      ~\protect\subref{sfig:lr-psnr} PSNR as a function of the map size in megabytes (\SI{}{\mega\byte})
      for each approach. The dataset under consideration is the synthetic D1 dataset shown in~\cref{sfig:lr-gt}.
      Note that the proposed approach yields a map that requires less disk
      space than the competing methods while demonstrating at par or better reconstruction accuracy
      (i.e., low reconstruction error and high precision).}}
  \vspace{-0.5cm}
\end{figure*}

\begin{figure*}
  \centering
  \ifthenelse{\equal{\arxivmode}{false}}
  {%
  \subfloat[\blue{Synthetic Dataset (D1)}\label{sfig:lr-gt}]{\includegraphics[width=0.2\textwidth,trim=820 390 770 370,clip]{figures/livingroom1-gt.eps}}
  \subfloat[Octomap (\SI{0.02}{\meter}, \SI{3.2}{\mega\byte})\label{sfig:lr-octo}]{\includegraphics[width=0.2\textwidth,trim=820 390 770 370,clip]{figures/livingroom1-octomap.eps}}
  \subfloat[\blue{Nvblox (\SI{0.08}{\meter}, \SI{6.4}{\mega\byte})}\label{sfig:lr-nvb}]{\includegraphics[width=0.2\textwidth,trim=820 390 770 370,clip]{figures/livingroom1-nvblox.eps}}
  \subfloat[\blue{FCGMM ($400$, \SI{4.8}{\mega\byte})}\label{sfig:lr-fcg}]{\includegraphics[width=0.2\textwidth,trim=820 390 770 370,clip]{figures/livingroom1-fcgmm.eps}}
  \subfloat[Proposed ($0.02$, \SI{2.1}{\mega\byte})\label{sfig:lr-ours}]{\includegraphics[width=0.2\textwidth,trim=820 390 770 370,clip]{figures/livingroom1-ours.eps}}\\
  \subfloat[\blue{Real-World Dataset (D2)}\label{sfig:lou-gt}]{\includegraphics[width=0.2\textwidth,trim=630 474 500 200,clip]{figures/lounge-gt.eps}}
  \subfloat[\blue{Octomap (\SI{0.02}{\meter}, \SI{5.9}{\mega\byte})}\label{sfig:lou-octo}]{\includegraphics[width=0.2\textwidth,trim=630 474 500 200,clip]{figures/lounge-octomap.eps}}
  \subfloat[\blue{Nvblox (\SI{0.06}{\meter}, \SI{8.2}{\mega\byte})}\label{sfig:lou-nvb}]{\includegraphics[width=0.2\textwidth,trim=630 474 500 200,clip]{figures/lounge-nvblox.eps}}
  \subfloat[\blue{FCGMM ($400$, \SI{6.2}{\mega\byte})}\label{sfig:lou-fcg}]{\includegraphics[width=0.2\textwidth,trim=630 474 500 200,clip]{figures/lounge-fcgmm.eps}}
  \subfloat[\blue{Proposed ($0.02$, \SI{5.7}{\mega\byte})}\label{sfig:lou-ours}]{\includegraphics[width=0.2\textwidth,trim=630 474 500 200,clip]{figures/lounge-isogmm.eps}}
  }%
  {%
  \subfloat[\blue{Synthetic Dataset (D1)}\label{sfig:lr-gt}]{\includegraphics[width=0.2\textwidth,trim=820 390 770 370,clip]{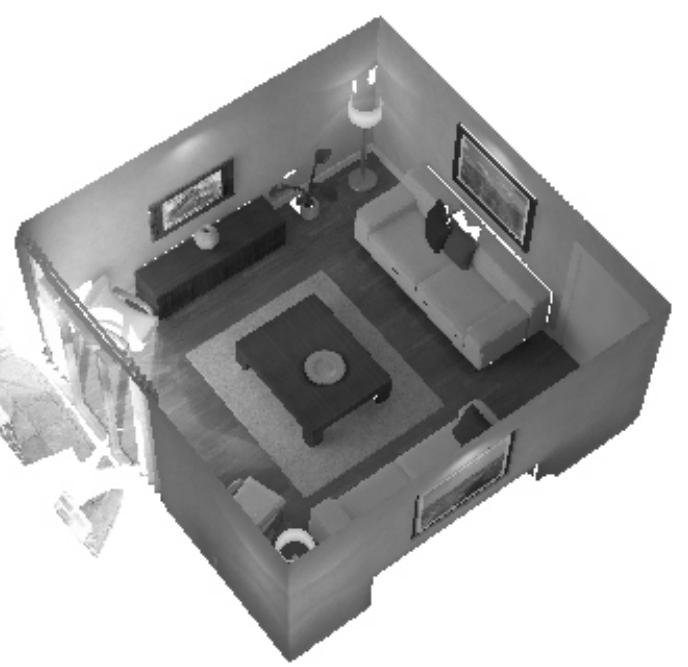}}
  \subfloat[Octomap (\SI{0.02}{\meter}, \SI{3.2}{\mega\byte})\label{sfig:lr-octo}]{\includegraphics[width=0.2\textwidth,trim=820 390 770 370,clip]{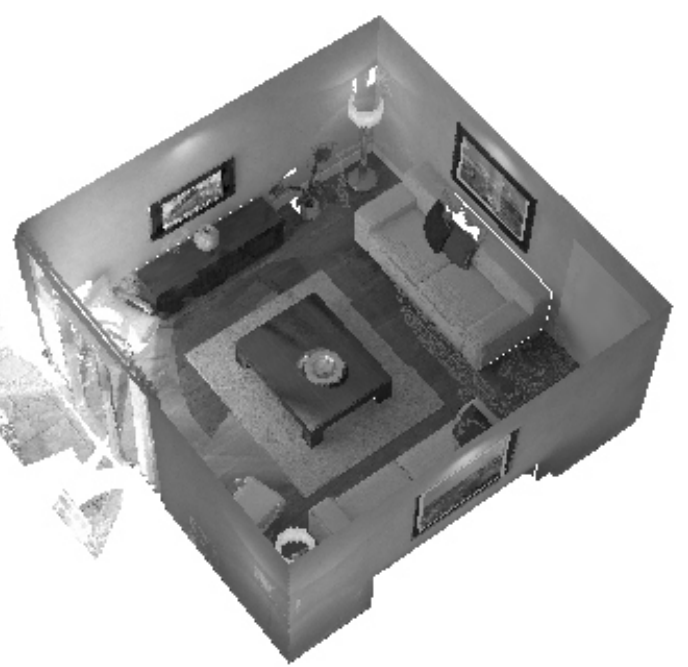}}
  \subfloat[\blue{Nvblox (\SI{0.08}{\meter}, \SI{6.4}{\mega\byte})}\label{sfig:lr-nvb}]{\includegraphics[width=0.2\textwidth,trim=820 390 770 370,clip]{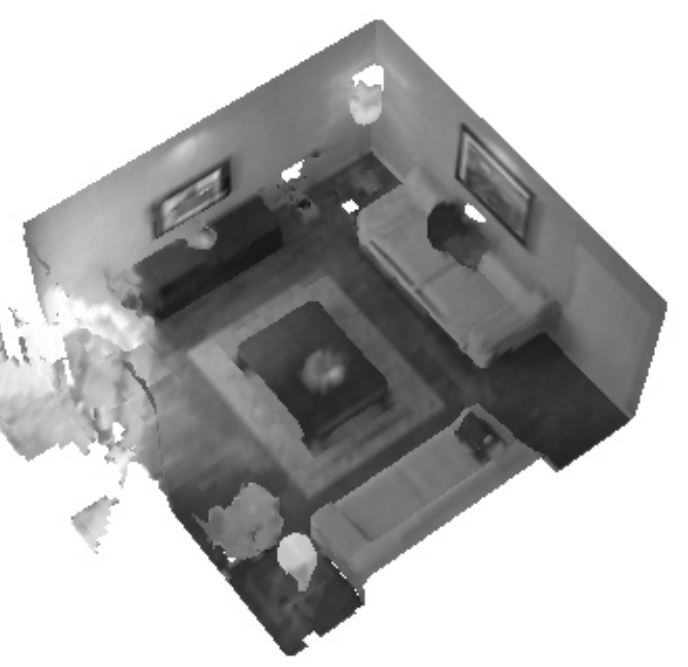}}
  \subfloat[\blue{FCGMM ($400$, \SI{4.8}{\mega\byte})}\label{sfig:lr-fcg}]{\includegraphics[width=0.2\textwidth,trim=820 390 770 370,clip]{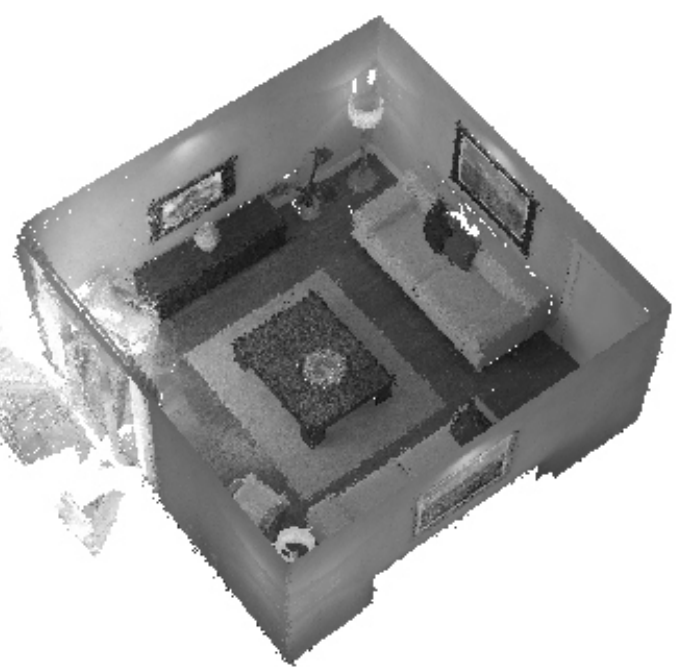}}
  \subfloat[Proposed ($0.02$, \SI{2.1}{\mega\byte})\label{sfig:lr-ours}]{\includegraphics[width=0.2\textwidth,trim=820 390 770 370,clip]{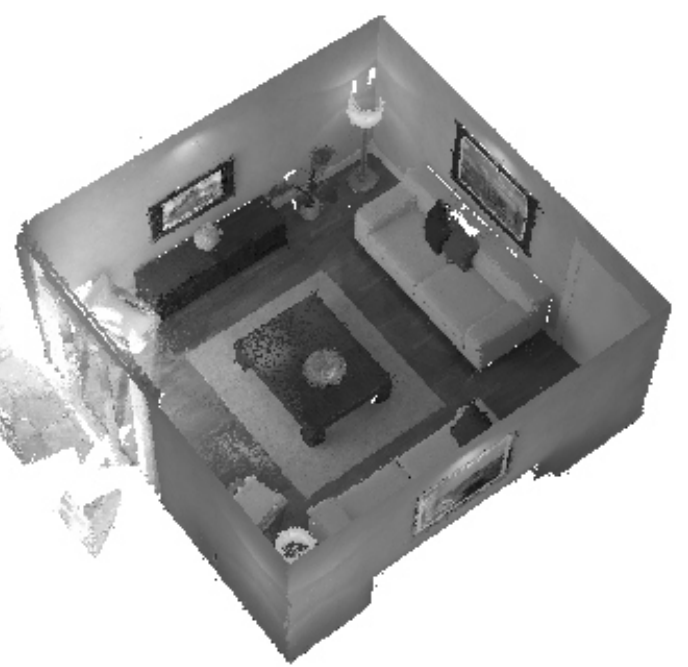}}\\
  \subfloat[\blue{Real-World Dataset (D2)}\label{sfig:lou-gt}]{\includegraphics[width=0.2\textwidth,trim=630 474 500 200,clip]{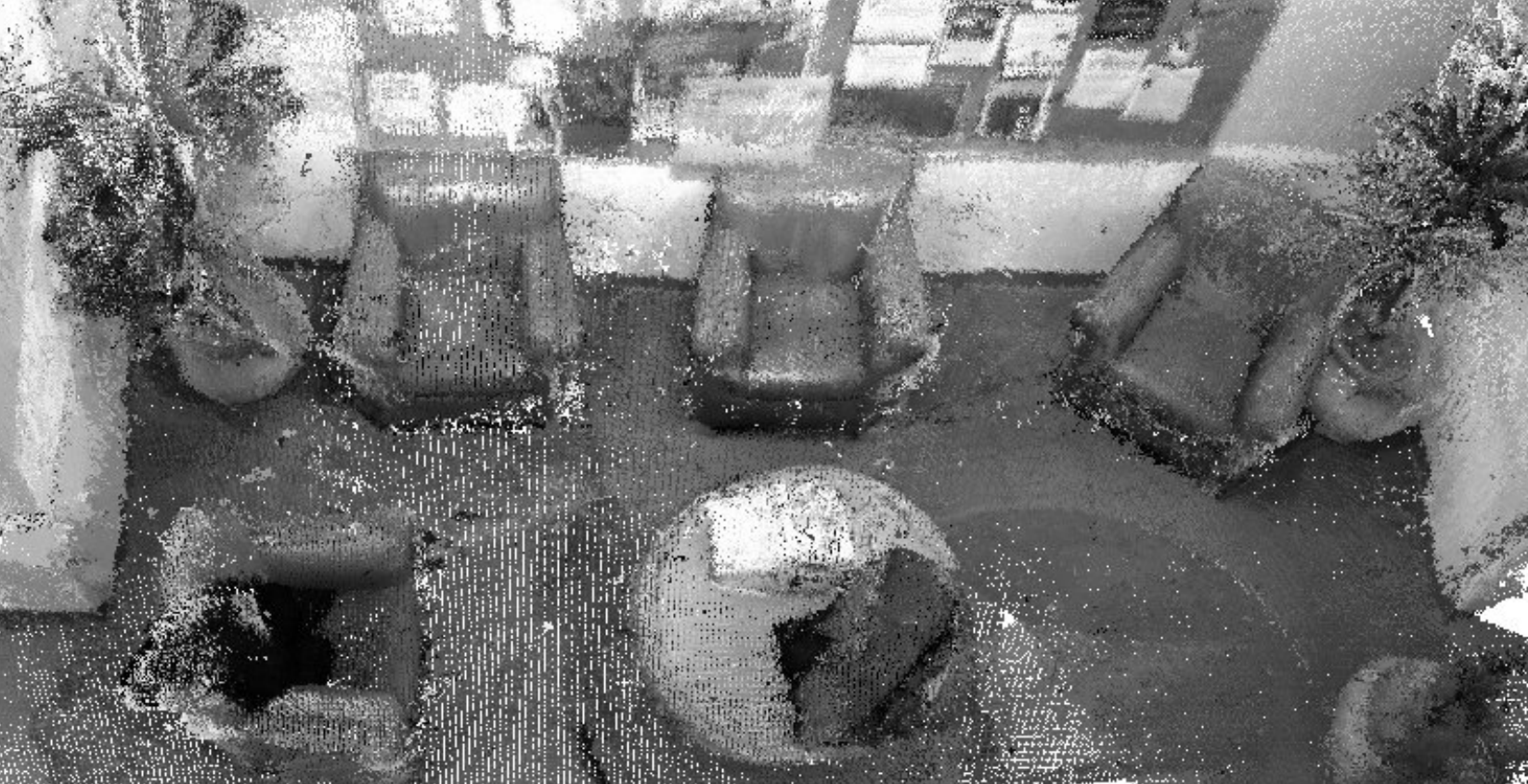}}
  \subfloat[\blue{Octomap (\SI{0.02}{\meter}, \SI{5.9}{\mega\byte})}\label{sfig:lou-octo}]{\includegraphics[width=0.2\textwidth,trim=630 474 500 200,clip]{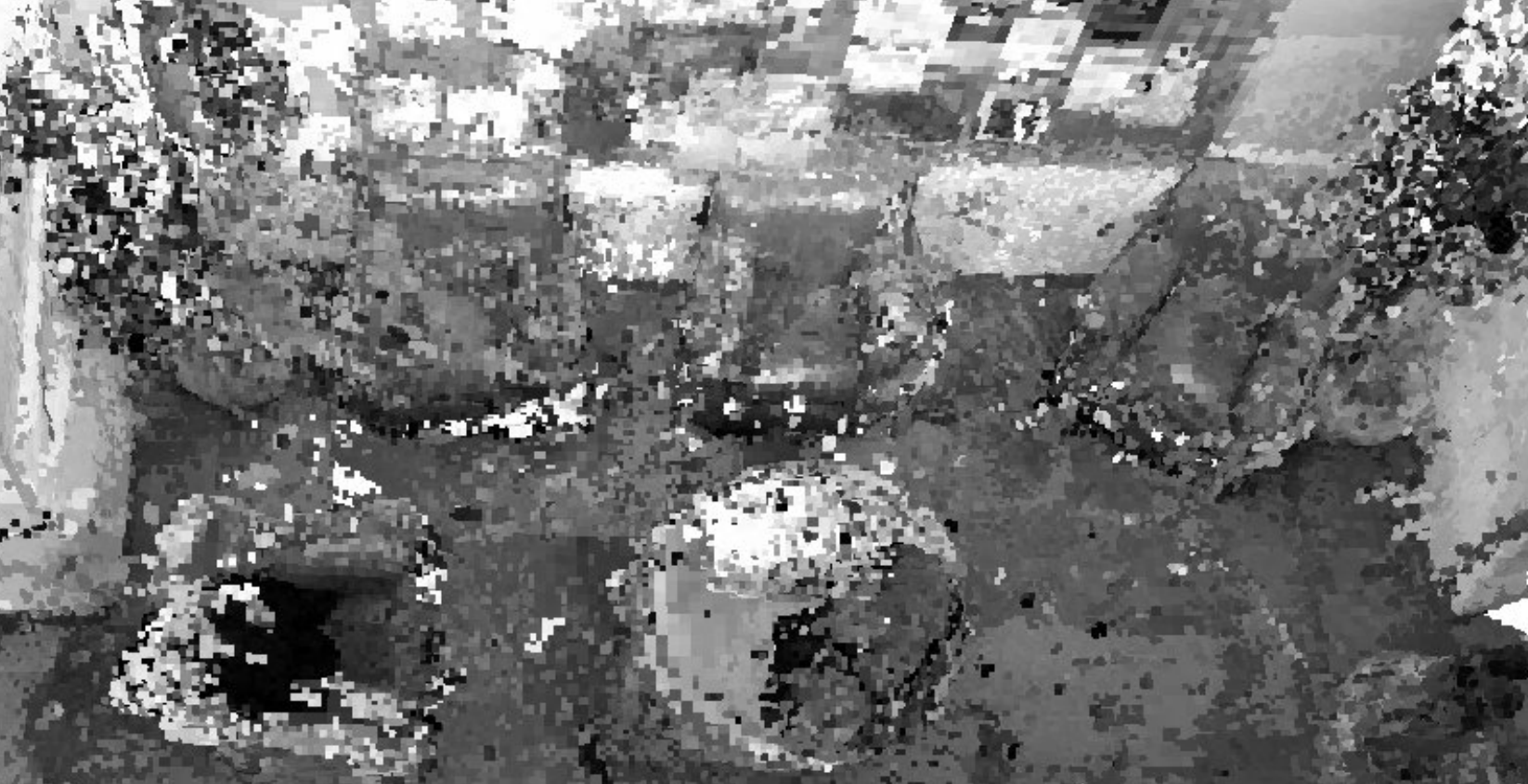}}
  \subfloat[\blue{Nvblox (\SI{0.06}{\meter}, \SI{8.2}{\mega\byte})}\label{sfig:lou-nvb}]{\includegraphics[width=0.2\textwidth,trim=630 474 500 200,clip]{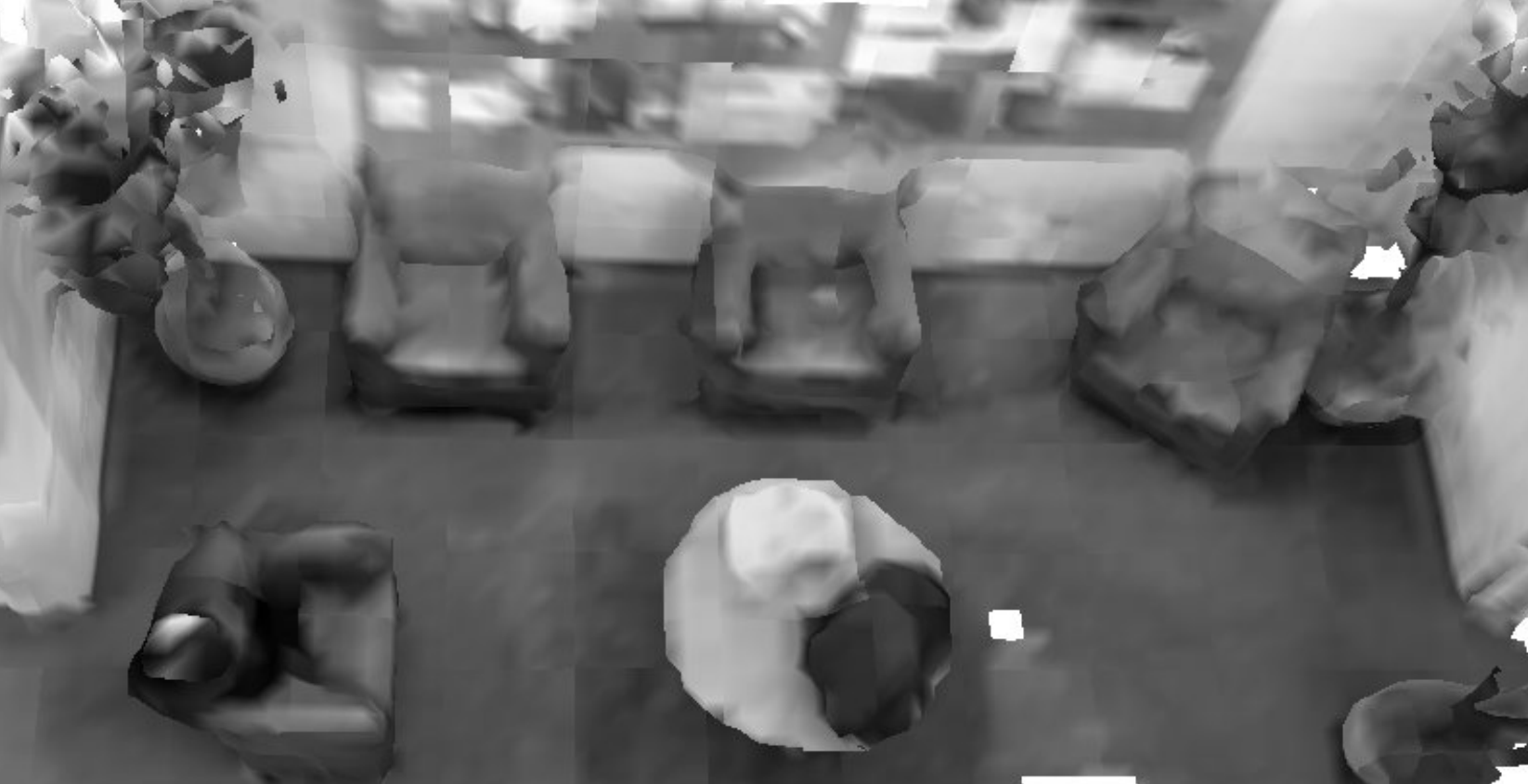}}
  \subfloat[\blue{FCGMM ($400$, \SI{6.2}{\mega\byte})}\label{sfig:lou-fcg}]{\includegraphics[width=0.2\textwidth,trim=630 474 500 200,clip]{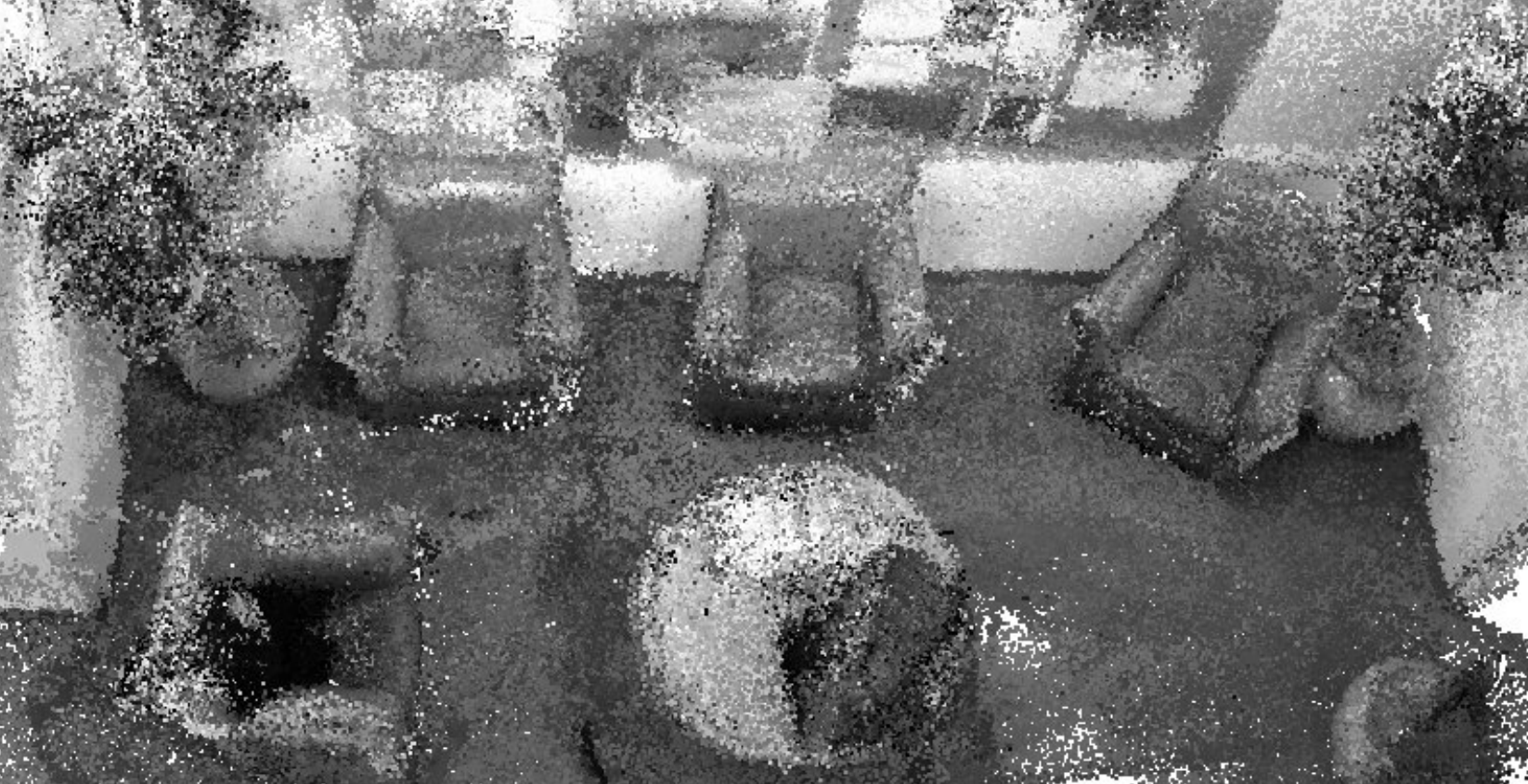}}
  \subfloat[\blue{Proposed ($0.02$, \SI{5.7}{\mega\byte})}\label{sfig:lou-ours}]{\includegraphics[width=0.2\textwidth,trim=630 474 500 200,clip]{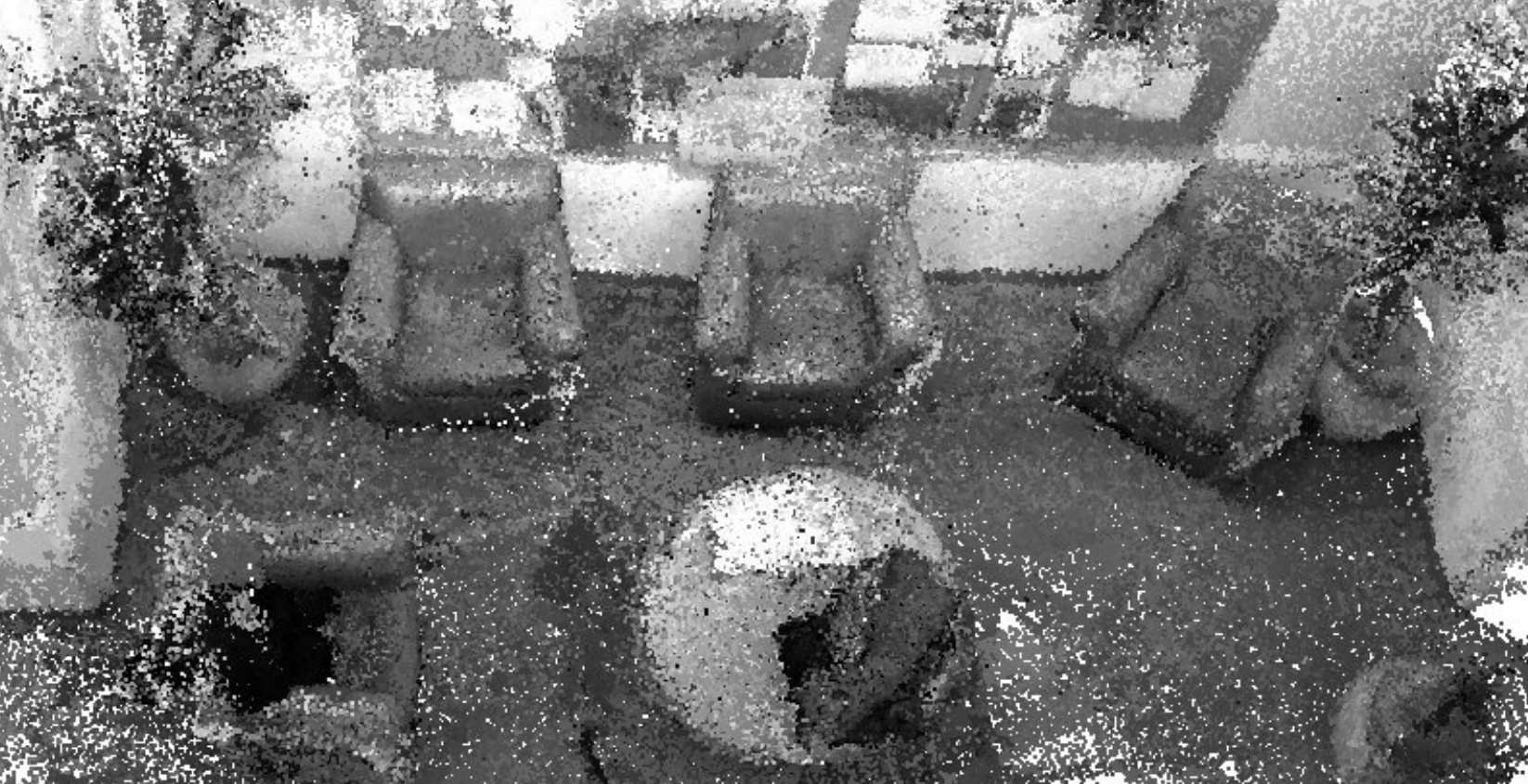}}
  }%
  \caption{\blue{\label{fig:results-qual}Qualitative comparison of the
  reconstructions obtained by baseline methods and the proposed approach at similar values of map size
  for~\protect\subref{sfig:lr-gt} D1 and~\protect\subref{sfig:lou-gt}
  D2 datasets. \wennie{The highest achievable resolution} used during execution and \wennie{resulting} map
  size\wennie{s} are reported in the sub-captions.
  ~\protect\subref{sfig:lr-octo} and~\protect\subref{sfig:lou-octo}
  \wennie{visualize the lowest level of the Octomap octree. I}ncorrect
  intensity values \wennie{are visible} due to the color averaging
  within the octree.
  ~\protect\subref{sfig:lr-nvb} and~\protect\subref{sfig:lou-nvb} \wennie{illustrate
  the mesh extracted from the stored TSDF for Nvblox. A}liasing \wennie{is visible}
  in the meshes due to large voxel sizes required for a lower memory footprint.
  ~\protect\subref{sfig:lr-fcg},~\protect\subref{sfig:lou-fcg} FCGMM and
  \protect\subref{sfig:lr-ours},~\protect\subref{sfig:lou-ours} the proposed method enable
  qualitatively similar high-resolution dense reconstructions; however, the
  FCGMM output requires a much longer time to process incremental observations
  (see~\cref{fig:ll-comparison}). A video of the proposed approach
  reconstructing the D1 dataset is available at
  \url{https://youtu.be/VgPEEcbUAnY}.}}
  \vspace{-0.5cm}
\end{figure*}

\begin{figure}
  \footnotesize
  \centering
  \subfloat[Performance measures for the D2: Lounge dataset\label{stab:lounge}]{\begin{tabular}{cc|cccc|c}
\toprule
Method & Param. & \shortstack{MRE\\(m) $\downarrow$} & Prec. $\uparrow$ & Rec. $\uparrow$ & \shortstack{PSNR\\(dB) $\uparrow$} & \shortstack{Mem.\\(MB) $\downarrow$} \\
\midrule
\multirow[t]{4}{*}{Octomap} & 0.02 & 0.007 & 0.81 & 0.46 & 18.37 & 5.94 \\
 & 0.04 & 0.011 & 0.55 & 0.06 & 16.76 & 1.13 \\
 & 0.06 & 0.016 & 0.41 & 0.02 & 15.84 & 0.44 \\
 & 0.08 & \red{\textbf{0.021}} & \red{\textbf{0.33}} & \red{\textbf{0.01}} & \red{\textbf{15.30}} & \green{\textbf{0.23}} \\
\hline
\multirow[t]{4}{*}{\blue{Nvblox}} & 0.02 & 0.006 & 0.92 & 0.38 & 20.63 & \red{\textbf{114.19}} \\
 & 0.04 & 0.008 & 0.86 & 0.35 & 19.08 & 19.98 \\
 & 0.06 & 0.011 & 0.81 & 0.31 & 18.26 & 8.19 \\
 & 0.08 & 0.014 & 0.76 & 0.28 & 17.51 & 4.22 \\
\hline
\multirow[t]{4}{*}{\blue{FCGMM}} & 800 & \green{\textbf{0.005}} & \green{\textbf{0.98}} & \green{\textbf{0.72}} & 20.25 & 14.11\\
 & 400 & 0.006 & 0.97 & \green{\textbf{0.72}} & 19.35 & 6.24 \\
 & 200 & \green{\textbf{0.005}} & 0.96 & \green{\textbf{0.72}} & 18.64 & 2.98 \\
 & 100 & 0.006 & 0.92 & \green{\textbf{0.72}} & 17.62 & 1.46 \\
\hline
\multirow[t]{4}{*}{Proposed} & 0.02 & \green{\textbf{0.005}} & \green{\textbf{0.98}} & 0.65 & \green{\textbf{20.74}} & 5.71 \\
 & 0.03 & \green{\textbf{0.005}} & 0.97 & 0.64 & 20.18 & 2.39 \\
 & 0.04 & 0.006 & 0.96 & 0.65 & 19.76 & 1.32 \\
 & 0.05 & 0.006 & 0.93 & 0.65 & 19.12 & 0.83 \\
\bottomrule
\end{tabular}}\\
  \subfloat[Performance measures for the D3: Copyroom dataset\label{stab:copyroom}]{\begin{tabular}{ccccccc}
\toprule
Method & Param. & \shortstack{MRE\\(m) $\downarrow$} & Prec. $\uparrow$ & Rec. $\uparrow$ & \shortstack{PSNR\\(dB) $\uparrow$} & \shortstack{Mem.\\(MB) $\downarrow$} \\
\midrule
\multirow[t]{4}{*}{Octomap} & 0.02 & 0.007 & 0.83 & 0.51 & 18.61 & 5.93 \\
 & 0.04 & 0.011 & 0.57 & 0.06 & 17.35 & 1.10 \\
 & 0.06 & 0.016 & 0.44 & 0.02 & 16.49 & 0.43 \\
 & 0.08 & \red{\textbf{0.021}} & \red{\textbf{0.36}} & \red{\textbf{0.01}} & \red{\textbf{15.95}} & \green{\textbf{0.23}} \\
\hline
\multirow[t]{4}{*}{\blue{Nvblox}} & 0.02 & 0.006 & 0.92 & 0.38 & 20.94 & \red{\textbf{96.95}} \\
 & 0.04 & 0.009 & 0.87 & 0.33 & 19.71 & 18.61 \\
 & 0.06 & 0.010 & 0.83 & 0.29 & 18.86 & 8.00 \\
 & 0.08 & 0.010 & 0.81 & 0.26 & 18.29 & 4.40 \\
\hline
\multirow[t]{3}{*}{\blue{FCGMM}} & 800 & -- & -- & -- & -- & -- \\
 & 400 & \green{\textbf{0.004}} & 0.97 & \green{\textbf{0.83}} & 20.66 & 10.68 \\
 & 200 & 0.005 & 0.95 & \green{\textbf{0.83}} & 19.70 & 4.85 \\
 & 100 & 0.007 & 0.91 & \green{\textbf{0.83}} & 18.59 & 2.30 \\
\hline
\multirow[t]{4}{*}{Proposed} & 0.02 & 0.005 & \green{\textbf{0.98}} & 0.70 & \green{\textbf{21.42}} & 5.59 \\
 & 0.03 & 0.005 & 0.97 & 0.68 & 20.96 & 2.48 \\
 & 0.04 & 0.005 & 0.95 & 0.66 & 20.56 & 1.39 \\
 & 0.05 & 0.006 & 0.93 & 0.66 & 20.05 & 0.88 \\
\bottomrule
\end{tabular}
}\\
  \subfloat[Performance measures for the D4: Stonewall dataset\label{stab:stonewall}]{\begin{tabular}{cc|cccc|c}
\toprule
Method & Param. & \shortstack{MRE\\(m) $\downarrow$} & Prec. $\uparrow$ & Rec. $\uparrow$ & \shortstack{PSNR\\(dB) $\uparrow$} & \shortstack{Mem.\\(MB) $\downarrow$} \\
\midrule
\multirow[t]{4}{*}{Octomap} & 0.02 & 0.007 & 0.84 & 0.62 & 19.19 & 5.29 \\
 & 0.04 & 0.011 & 0.59 & 0.08 & 18.32 & 0.95 \\
 & 0.06 & 0.016 & 0.47 & 0.02 & 17.61 & 0.37 \\
 & 0.08 & \red{\textbf{0.021}} & \red{\textbf{0.39}} & \red{\textbf{0.01}} & \red{\textbf{17.79}} & \green{\textbf{0.19}} \\
\hline
\multirow[t]{4}{*}{\blue{Nvblox}} & 0.02 & 0.005 & \green{\textbf{0.99}} & 0.42 & \green{\textbf{24.80}} & \red{\textbf{155.05}} \\
 & 0.04 & 0.005 & 0.98 & 0.39 & 23.41 & 25.96 \\
 & 0.06 & 0.005 & 0.96 & 0.37 & 22.40 & 9.43 \\
 & 0.08 & 0.006 & 0.94 & 0.34 & 21.43 & 4.89 \\
\hline
\multirow[t]{4}{*}{\blue{FCGMM}} & 800 & 0.005 & \green{\textbf{0.99}} & \green{\textbf{0.82}} & 22.13 & 13.34 \\
 & 400 & \green{\textbf{0.004}} & \green{\textbf{0.99}} & 0.80 & 21.51 & 5.66 \\
 & 200 & 0.005 & 0.98 & 0.79 & 20.99 & 2.60 \\
 & 100 & 0.006 & 0.96 & 0.79 & 20.40 & 1.24 \\
\hline
\multirow[t]{4}{*}{Proposed} & 0.02 & 0.005 & \green{\textbf{0.99}} & 0.68 & 21.83 & 3.27 \\
 & 0.03 & 0.005 & 0.98 & 0.66 & 21.62 & 1.26 \\
 & 0.04 & 0.005 & 0.97 & 0.65 & 21.41 & 0.63 \\
 & 0.05 & 0.006 & 0.95 & 0.65 & 21.16 & 0.36 \\
\bottomrule
\end{tabular}
}
  \caption{
    \label{fig:real-quant}Quantitative comparison of Octomap, Nvblox, \blue{FCGMM}, and the
    proposed approach using the real-world datasets with noisy RGB-D data.
    \blue{The best and worst values in each column are colored green and red
    respectively.} \blue{The FCGMM method results in a larger map size compared to the
    proposed approach and is orders of magnitude slower in execution time
    (\cref{fig:ll-comparison}).} These results highlight that the proposed
    approach balances the accuracy and \blue{map size} better than the
    state-of-the-art approaches.}
\end{figure}

One synthetic (D1: Living Room from the Augmented ICL-NUIM
datasets~\citep{choi_robust_2015}) and three real-world datasets (D2: Lounge,
D3: Copyroom, and D4: Stonewall from the Redwood
datasets~\citep{zhou_dense_2013}) are used for qualitative and quantitative
evaluation of the proposed approach and its comparison with the baseline
methods. All datasets contain \blue{$640 \times 480$} RGB and depth images along
with the corresponding camera poses. \blue{All methods are provided reduced
resolution $128 \times 96$ images for incremental mapping. The computer used for
all experiments contains an Intel Core i9-10900K CPU (20 threads, maximum clock
speed \SI{5.3}{\giga\hertz}, \SI{32}{\giga\byte} RAM) and a NVIDIA GeForce RTX
3060 GPU (\SI{12}{\giga\byte} RAM). The CPU implementation of the
Octomap approach for color data is used. Nvblox and the proposed
method\footnote{Release~\texttt{0.1.0} of~\url{https://github.com/gira3d/gira3d-reconstruction}.} use the CPU
and GPU for incremental mapping.} Nvblox uses both depth and color data by default. \blue{It is modified to use
depth and grayscale images for the comparison presented in this section.}
\blue{Since the software for prior GMM map
works~\citep{srivastava_efficient_2017,srivastava_efficient_2019,tabib_autonomous_2021}
is not openly available, the codebase for the proposed approach is modified to
use a fixed number of components for the FCGMM comparison. The FCGMM approach
uses the GPU for EM execution but CPU for \wennie{the} $\pointcloud^r$ \wennie{calculation} because it requires
access to \wennie{more RAM than is available to the GPU}.}

\blue{\subsection{Relevant Point Cloud Calculation\label{ssec:ll-compare}}
As mentioned in~\Cref{sssec:local-sogmm}, a drawback of prior work~\citep{srivastava_efficient_2017}
is that the relevant point cloud subset $\pointcloud^r$ calculation
in~\cref{eq:zr-4d} \wennie{is not real-time viable, especially} as the number of
components $|\gindexset|$ increase in the global point cloud model $\gsogmm$.
The proposed spatial hashing approach reduces this computation cost by selecting a subset
of components $\findexset$ that geometrically overlaps the point cloud.~\Cref{fig:ll-comparison}
demonstrates the performance gains due to the proposed approach.
While the methodology proposed in~\citep{srivastava_efficient_2017} is
hierarchical, we choose to compare against their highest fidelity model (i.e.,
lowest layer) in the hierarchy for a fair comparison.

For the baseline approach (\cref{sfig:ll-base}), the calculation times
per frame are shown for increasing values of number of mixture
components $|\lindexset|$, $\leaffcgmm = \{ 100, 200, 400, 800
\}$. \wennie{The proposed methodology enables an order of magnitude
faster $\pointcloud^r$ calculation as compared to the baseline
approach, because the log-likelihood operates over all mixture
components and points.}  \wennie{T}he spatial hash resolution $\alpha$
is fixed at $\SI{0.2}{\meter}$ \wennie{for all values of
$\bandwidth$}. \wennie{Because lower $\bandwidth$ yields higher
resolution reconstruction, the computation time increases as
$\bandwidth$ decreases.}
~\Cref{sfig:ll-times} presents the cumulative $\pointcloud^r$ calculation
times for the D1 dataset. The $\Delta$ columns show the order of magnitude
improvement via the proposed approach. Notice that while the performance gain of
Proposed-$0.05$ compared to FCGMM-$100$ is nearly $\mathbf{10\times}$, it increases
with the model fidelity; the Proposed-$0.02$ is about $\mathbf{30\times}$
faster than FCGMM-$800$. Finally,~\cref{sfig:ll-sp-abl} compares the $\pointcloud^r$
calculation times for $\mres = \{ 0.1, 0.2, 0.4, 0.8 \}$. Increasing $\mres$ results
in overall higher calculation times because the size of $\findexset$ gets larger.
Before using the proposed method, the value of $\mres$ should be set according to the available
CPU computation resources.
}

\blue{\subsection{Global Map Accuracy and Compression\label{ssec:gl-map-compare}}}
For Octomap and Nvblox, the predicted point cloud $\prpcld$ \blue{and mesh are}
constructed\wennie{,} \blue{respectively}\wennie{,} after processing all the frames in a given
dataset. For \blue{FCGMM} and the proposed approach, the predicted point cloud is
inferred from the global model $\gsogmm$ using the method in~\cref{ssec:proposed-inference}.
The Octomap method requires specifying a minimum leaf size for the underlying
octree used for modeling and inference. We use a range of leaf sizes for the
experiments, $\leafom = \{ 0.02, 0.04, 0.06, 0.08 \} \SI{}{\meter}$. The same
set of values are used for the voxel sizes required in the Nvblox method,
$\leafnv = \leafom$. \blue{For FCGMM and the proposed approach, \wennie{the} same set of
parameters are used as in~\cref{ssec:ll-compare}.} The ground truth point cloud,
$\gtpcld$, is constructed by appending all the point clouds corresponding to
the images and poses in the dataset followed by downsampling using a voxel grid filter with a
small voxel size (\SI{0.01}{\meter} for all experiments in this section).

The performance measures for 3D reconstruction are (1) Mean \wennie{R}econstruction \wennie{E}rror
(MRE), which is the average distance of the closest points between $\prpcld$ and
$\gtpcld$ (lower is better), (2) Precision of 3D reconstruction, which measures
the fraction of points in $\prpcld$ that lie within \SI{0.01}{\meter} of a point
in $\gtpcld$ (higher is better), and (3) Recall of 3D reconstruction, which
measures the fraction of points in $\gtpcld$ that lie within \SI{0.01}{\meter}
of a point in $\prpcld$ (higher is better). \blue{Intuitively, this measure computes
the degree of ``completeness'' in the reconstruction.} The performance measure
for intensity reconstruction is the peak-signal-to-noise ratio (PSNR) calculated
using the mean squared error (MSE) between the intensity values of the closest
points in $\prpcld$ and $\gtpcld$ (higher PSNR is better).  \blue{These measures
are computed using the closest point distance computation functions for point
clouds in Open3D~\citep{zhou_open3d_2018}. For the Nvblox case in particular,
the mesh output is uniformly and densely sampled to create a point cloud with
number of points equal to the ground truth point cloud.}

The memory \blue{storage} efficiency of the multimodal environment
representations is measured by calculating the size (measured in bytes) of the
models that can be loaded from disk to create $\prpcld$. \blue{The models are
chosen so that they can enable reconstruction of the surface and occupancy
modeling for other robots in a multi-robot exploration scenario.} For Octomap,
the model size corresponds to the output \texttt{.ot}
file~\citep{hornung_octomap_2013}, \blue{as opposed to the binary \texttt{.bt}
file that does not retain occupancy information}. \blue{Due to the same reason,}
for Nvblox the SQLite3 database (\texttt{.db} file) output is used \blue{instead
of the output \texttt{.ply} mesh file}. For the \blue{FCGMM} and proposed
\blue{methodologies}, the memory occupied by the means, covariances, and weights
in the global model $\gsogmm$ is calculated (four floats for each mean, one
float for each weight, and ten floats for each covariance). \blue{Note that
occupancy modeling from a stored GMM map with these parameters has been
demonstrated in prior
work~\citep{eckart_accelerated_2016,srivastava_efficient_2019,omeadhra_variable_2019,tabib_autonomous_2021}}

\Cref{fig:livingroom1-plot} shows the variation in the performance measures for
all \blue{methods} with respect to the \blue{map size on disk} for the D1
dataset. Each data point in the plots corresponds to a unique parameter setting
in $\leafom$, $\leafnv$, \blue{$\leaffcgmm$}, and $\leafsogmm$ for Octomap,
Nvblox, \blue{FCGMM}, and the proposed approach respectively. To attain similar
levels of \blue{mean reconstruction error and precision}, the proposed
approach\blue{, FCGMM,} and Octomap require an order of magnitude \blue{less
memory than Nvblox (\cref{sfig:lr-mre,sfig:lr-prec})}.  This is
because Nvblox utilizes a regular grid of fixed resolution \blue{and multiple
data storage layers} while Octomap\blue{, FCGMM,} and the proposed approach
leverage octrees, GMMs\blue{, and SOGMMs}, respectively. \blue{Note that while
these values are close for the FCGMM and the proposed methods, in the FCGMM case
the time taken to create the model is much higher (\cref{ssec:ll-compare}).}

\blue{The FCGMM and the proposed approaches
achieve a recall score close to $1.0$ (\cref{sfig:lr-rec}) demonstrating that for nearly each point
in the ground truth point cloud, there is a point in the reconstruction within a
$\SI{0.01}{\meter}$ ball. Both of these GMM-based methods outperform Octomap
because an arbitrarily high number of points can be densely sampled from a GMM
(\cref{ssec:proposed-inference}) whereas the Octomap outputs the point cloud
at its minimum pre-specified leaf size. \wennie{The Nvblox method output
mesh is uniformly sampled; however, a low voxel size is required to achieve
similar recall scores.}}

The highest intensity reconstruction accuracy (i.e., PSNR score
in~\cref{sfig:lr-psnr}) attained by Octomap ($\text{PSNR} = 26.70$) is much
lower than the proposed approach ($\text{PSNR} = 30.29$) \blue{at a similar
storage cost}. This is because the intensity in an Octomap octree node is
averaged according to the density of the points around the node.  In contrast,
the proposed approach \wennie{treats} intensity \wennie{as} a univariate random variable and
jointly modeled with the 3D coordinates in the global point cloud model.
Inference from this joint probability density leads to a higher accuracy than
the averaging in Octomap. \blue{Nvblox fuses intensity information
into the 3D map using a weighted average update. This process is an improvement
over Octomap but still requires a low voxel size to attain a PSNR comparable to
GMM-based approaches. Finally, the FCGMM approach demonstrates a lower PSNR than
the proposed approach for similar storage costs. This is because the FCGMM uses
a fixed number of components for every scene in the dataset while the proposed method
uses SOGMM\wennie{, which} adapts the number of components according to the complexity of depth
and image data~\citep{goel_probabilistic_2023}.}
The impact of these quantitative results is visible in the qualitative
comparison shown in~\cref{fig:results-qual}. The reconstructions from all
methods are shown for comparable map sizes along with the ground truth point
cloud \blue{for D1 and D2 datasets}.

\Cref{fig:real-quant} provides performance statistics corresponding to
real-world datasets D2, D3 and D4, which exhibit noisy sensor readings.
\blue{For each performance measure, the best and worst values are
highlighted in green and red\wennie{,} respectively. Note that there is no result for the
D2 dataset in the FCGMM-$800$ case because the relevant point cloud calculation
required more RAM than the available $\SI{32}{\giga\byte}$. This is expected
since the D2 dataset contains $5490$ frames which is nearly twice the other
datasets. While using Octomap at a $\SI{0.08}{\meter}$ resolution results in the lowest
map size ($\SI{0.23}{\mega\byte}$), reconstruction performance is poor compared
to all other methods. Nvblox outputs the largest map size on disk at
$\SI{114.19}{\mega\byte}$ but does not enable the highest PSNR. The FCGMM and
proposed approach enable similar reconstruction accuracy; however, the proposed
approach results in smaller map sizes and utilizes \wennie{less} computation
as shown earlier. This trend is observed for the D2 and D3 datasets as well.
One exception for Nvblox is that it provides a higher PSNR compared to the
proposed approach for the D3 dataset but it consumes about $50 \times$ more
storage.
}


\blue{\section{Limitations\label{sec:limitations}}
The effects of changes in illumination in the scene are not explicitly
considered and requires future work. The proposed method assumes drift-free
sensor poses are available through a localization system. This assumption is
consistent with prior work on multimodal mapping with
GMMs~\citep{srivastava_efficient_2017}. Prior work in sensor localization via
point cloud registration~\citep{eckart_hgmr_2018,tabib_-manifold_2018} and loop
closure~\citep{tabib_simultaneous_2021} using GMMs may be leveraged for
pose estimation.
}

\section{Conclusion\label{sec:conclusion}}
This letter detailed an incremental multimodal surface mapping
methodology \wennie{for} high-resolution environment
reconstruction. \blue{State-of-the-art} \wennie{GMM-based} perceptual
modeling approaches use a pre-specified number of components to enable
mapping of environment surfaces, which \wennie{is} memory
inefficien\wennie{t}. \wennie{I}nserting a new point cloud observation
to an existing GMM map model involves iterating over all the mixture
components; which is computationally expensive.  To bridge these gaps,
this paper \wennie{formulated methodologies to (1)
extract a submap by innovating a spatial hash
table of mixture components and (2) incrementally update the global environment model in a computationally efficient manner.}
\wennie{The approach was evaluated with synthetic and real-world
datasets and the results demonstrated that the proposed approach
enables high-fidelity reconstruction at low memory with an order of
magnitude increase in speed compared to existing GMM-based mapping methods.}


{
  \footnotesize
  \balance
  \bibliographystyle{IEEEtranN}
  \bibliography{refs,content/bibliography/kshitij_library_without_urls}
}

\end{document}